\documentclass[twocolumn,iicol]{svjour3}

\usepackage{graphicx}%
\usepackage{multirow}%
\usepackage{amsmath,amssymb,amsfonts}%
\usepackage{mathrsfs}%
\usepackage[title]{appendix}%
\usepackage{xcolor}%
\usepackage{textcomp}%
\usepackage{manyfoot}%
\usepackage{booktabs}%
\usepackage{algorithm}%
\usepackage{algorithmic}
\usepackage{listings}%
\usepackage{adjustbox}
\usepackage{natbib}
\usepackage{times}

\usepackage[T1]{fontenc}
\usepackage{subcaption}
\usepackage{placeins}
\usepackage{pifont}
\usepackage[colorlinks,linkcolor=blue,citecolor=blue]{hyperref}
\usepackage{tabularray}
\usepackage[misc]{ifsym}
\usepackage{hyperref}
\usepackage[colorlinks,linkcolor=blue,citecolor=blue]{hyperref}

\usepackage{xurl}

\usepackage{enumitem}
\definecolor{mygray}{gray}{.9}
\usepackage[table,xcdraw]{xcolor}

\usepackage{subcaption}
\usepackage{makecell}

\usepackage[algo2e]{algorithm2e}

\begin{document}
\begin{sloppypar}
\title{AmPLe: Supporting Vision-Language Models via Adaptive-Debiased Ensemble Multi-Prompt Learning}



 \author{Fei Song\textsuperscript{\rm 1 \rm 2}$^{\ast}$ \and Yi Li\textsuperscript{\rm 1 \rm 2}$^{\ast}$ \and Jiangmeng Li\textsuperscript{\rm 1}$^{\ast}$ \and Rui Wang\textsuperscript{\rm 1 \rm 2} \and Changwen Zheng\textsuperscript{\rm 1 \rm 2} \and Fanjiang Xu\textsuperscript{\rm 1 \rm 2} \and Hui Xiong\textsuperscript{\rm 3 \rm 4}}
 

\institute{$^{\ast}$ Equal contribution. \\
            \textrm{\Letter} Jiangmeng Li \\
            \email{jiangmeng2019@iscas.ac.cn} \\
            {$^1$} National Key Laboratory of Space Integrated Information System, Institute of Software, Chinese Academy of Sciences, Beijing, China \\
            {$^2$} University of Chinese Academy of Sciences, Beijing, China \\
            {$^3$} The Hong Kong University of Science and Technology (Guangzhou), Guangdong, China\\
            {$^4$} The Hong Kong University of Science and Technology, Hong Kong SAR, China\\
 }

\date{Received: date / Accepted: date}
\maketitle
\begin{abstract}
Multi-prompt learning methods have emerged as an effective approach for facilitating the rapid adaptation of vision-language models to downstream tasks with limited resources. Existing multi-prompt learning methods primarily focus on utilizing various meticulously designed prompts within a single foundation vision-language model to achieve superior performance. However, the overlooked \textit{model-prompt matching bias} hinders the development of multi-prompt learning, i.e., the same prompt can convey different semantics across distinct vision-language models, such as CLIP-ViT-B/16 and CLIP-ViT-B/32, resulting in inconsistent predictions of identical prompt. To mitigate the impact of this bias on downstream tasks, we explore an ensemble learning approach to sufficiently aggregate the benefits of diverse predictions. Additionally, we further disclose the presence of \textit{sample-prompt matching bias}, which originates from the prompt-irrelevant semantics encapsulated in the input samples. Thus, directly utilizing all information from the input samples for generating weights of ensemble learning can lead to suboptimal performance. In response, we extract prompt-relevant semantics from input samples by leveraging the guidance of the information theory-based analysis, adaptively calculating debiased ensemble weights. Overall, we propose Adaptive-Debiased Ensemble Multi-Prompt Learning, abbreviated as AmPLe, to mitigate the two types of bias simultaneously. Extensive experiments on three representative tasks, i.e., generalization to novel classes, new target datasets, and unseen domain shifts, show that AmPLe can widely outperform existing methods. Theoretical validation from a causal perspective further supports the effectiveness of AmPLe. The source code can be accessed at \href{https://github.com/FF2127/AmPLe}{AmPLe}.

\end{abstract}

\section{Introduction}

In the past few years, methods~\citep{DBLP:journals/ijcv/ZhouYLL22, DBLP:conf/cvpr/ZhouYL022, DBLP:conf/cvpr/KhattakR0KK23, DBLP:conf/iccv/KhattakWNK0K23, DBLP:conf/eccv/MirzaKLDMKKP24} that combine prompt learning~\citep{DBLP:conf/nips/BrownMRSKDNSSAA20} with foundation vision-language models (VLMs), such as Contrastive Language-Image Pretraining (CLIP)~\citep{DBLP:conf/icml/RadfordKHRGASAM21}, have demonstrated excellent generalization ability across various domains. By incorporating learnable vectors into the textual prompt and conducting alignment between visual and textual features, these methods enable VLMs to effectively adapt to diverse tasks. This not only enhances the performance of VLMs across a wide range of downstream tasks~\citep{DBLP:conf/cvpr/BangA024, DBLP:conf/icml/NasirianyX0XL0X24, DBLP:conf/cvpr/DuanCXWZYX22} but also strengthens the VLMs' robustness in cross-domain scenarios~\citep{DBLP:conf/nips/Li0YSHZLC23, DBLP:journals/tip/ZhaoWJSSLM24, DBLP:conf/aaai/BaiZZHLWC24}, particularly under the scenarios of limited resources and sparse labeled data.

\begin{figure}
    \centering
    \includegraphics[width=0.48\textwidth]{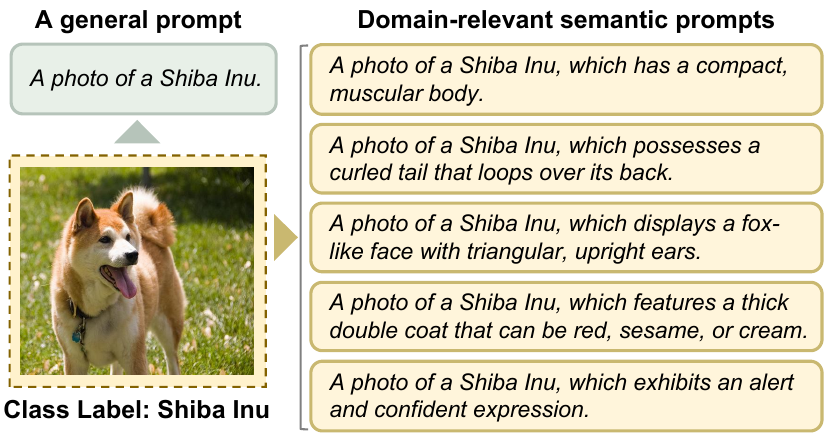}
    \caption{An example of using multiple prompts to describe the Shiba Inu class.}
    \label{fig:multi_prompts}
\end{figure}

Although prompt learning has documented promising results~\citep{DBLP:journals/corr/abs-2309-04158,DBLP:conf/eccv/WuZLCLYL24,DBLP:journals/ijcv/BulatT24,yang2025learning} in effectively adapting foundation VLMs to downstream tasks, a single textual prompt is intuitively insufficient to fully capture the essence of a class.
A class is typically represented through multiple visual descriptors, each providing complementary fine-grained portrayal.
As shown in Fig. \ref{fig:multi_prompts}, when describing the class ``Shiba Inu'', we can use a simple \textit{general prompt}, such as ``A photo of a Shiba Inu.'', or employ prompts that describe the class from different perspectives, such as its ``body'', ``tail'', ``face'', ``ears'', and so on. These various prompts capture the basic characteristics of the Shiba Inu by providing comprehensive descriptions of its appearance, behavior, and other distinguishing features. We refer to these prompts, which incorporate different class-relevant descriptors, as \textit{domain-relevant semantic prompts}. By combining the general prompt and domain-relevant semantic prompts, the multiple prompts can describe the ``Shiba Inu'' class from both coarse-grained and fine-grained levels, thereby offering a more comprehensive and precise representation of its characteristics. 

\begin{figure}
    \centering
    \includegraphics[width=0.48\textwidth]{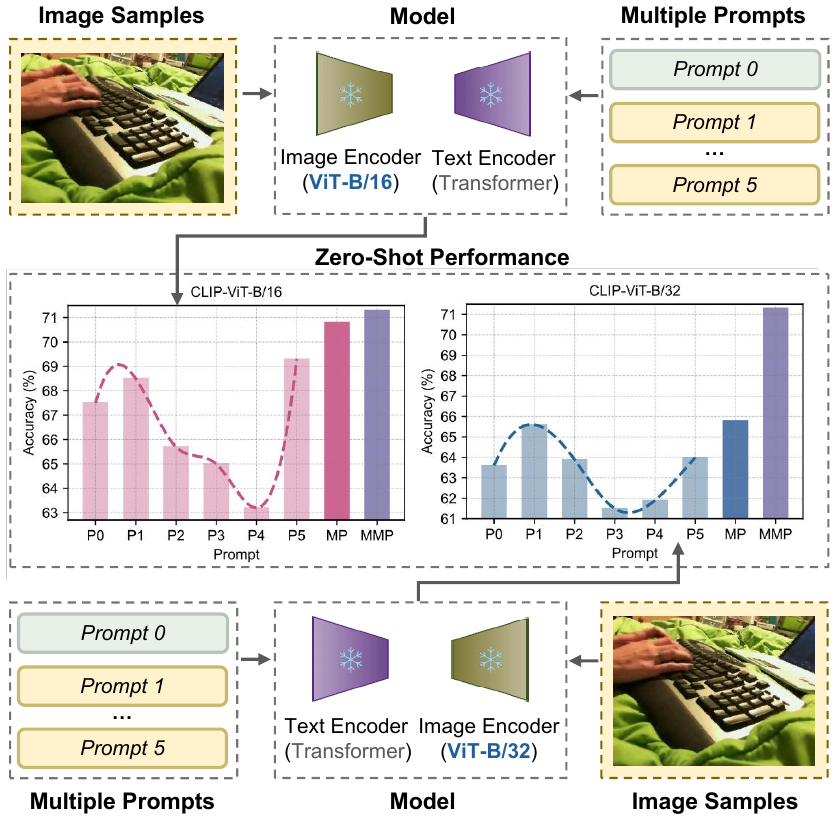}
    \caption{Zero-shot performance using multiple prompts and different CLIP models on the UCF101 dataset. In the histograms, P0 to P5 correspond to predictions obtained by applying a single prompt in a single VLM (CLIP-ViT-B/16 or CLIP-ViT-B/32), MP is the simple aggregation (sum and average) of P0 to P5's diverse predictions within the same VLM, and MMP aggregates the diverse predictions from P0 to P5 across both CLIP-ViT-B/16 and CLIP-ViT-B/32. The best result of MMP highlights the potential of combining multiple prompts and different VLMs to improve model performance.}
    \label{fig:motivation1}
\end{figure}

However, existing multi-prompt learning methods~\citep{DBLP:conf/miccai/GuoYQWML23, DBLP:conf/iclr/0002YSLR023, DBLP:conf/iccv/ChenHGL023, DBLP:conf/mir/LiuSC24} primarily focus on applying multiple prompts within a single VLM to achieve superior performance, overlooking the
existence of \textit{model-prompt matching bias}. This bias indicates that the identical prompt can exhibit distinct semantic information to different VLMs, leading to diverse predictions corresponding to the same prompt. To substantiate this claim, we investigate the zero-shot performance of two widely adopted CLIP models, i.e., CLIP-ViT-B/16 and CLIP-ViT-B/32, on the UCF101~\citep{DBLP:journals/corr/abs-1212-0402} dataset. 
As shown in Fig. \ref{fig:motivation1}, when evaluating zero-shot performance with the CLIP-ViT-B/16 model, the ``P5'' prompt performs the best among six individual prompts, while the ``P1'' prompt yields the highest performance with the CLIP-ViT-B/32 model. The results confirm that different VLMs extract different semantic information from the input image under the guidance of identical prompts. Accordingly, inspired by the preeminent performance of MP and MMP in Fig. \ref{fig:motivation1} and \citep{DBLP:conf/icml/Qiao024}, we proceed to aggregate the diverse prompt-specific predictions across different VLMs based on the principles of ensemble learning~\citep{DBLP:journals/widm/SagiR18}.

\begin{figure}
    \centering
    \includegraphics[width=0.48\textwidth]{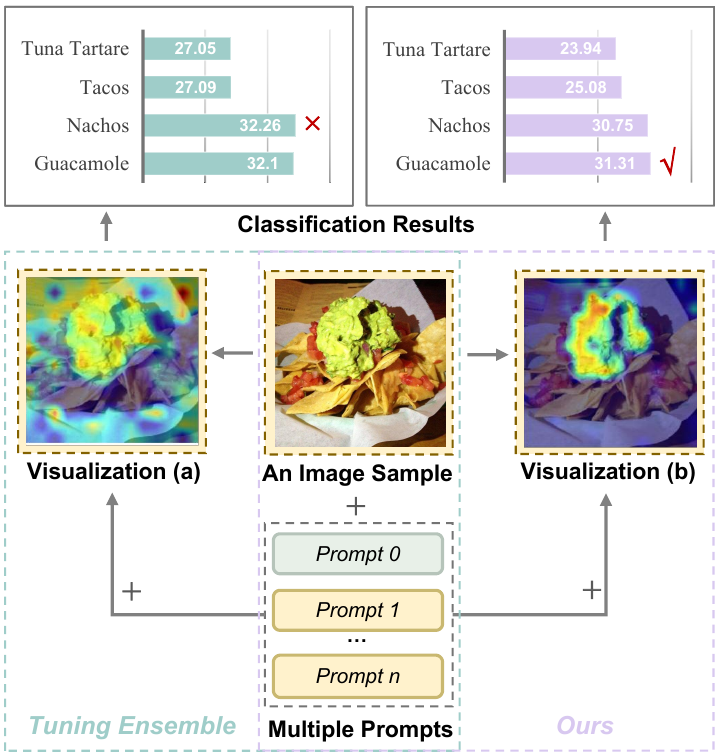}
    \caption{Comparison of attention maps for the Guacamole class in the Food101 dataset. Visualization (a) shows the attention map using Tuning Ensemble for image classification, demonstrating that the model attends to both the Guacamole object and irrelevant areas when using all information of the input image for ensemble weights generation. Visualization (b) shows the attention map when using prompt-relevant information from the input image for ensemble weights generation, illustrating that the model focuses primarily on the Guacamole object with minimal attention to irrelevant areas.}
    \label{fig:motivation2}
\end{figure}

The calculation of ensemble weights is crucial for achieving significant performance gains. The state-of-the-art method, Tuning Ensemble \citep{DBLP:conf/icml/LuB0XW24}, adopts a dynamic strategy in which the ensemble weights are determined by feeding the features of input samples into a neural network-based weight generator. Nevertheless, another bias hinders the accurate determination of ensemble weights, namely, \textit{sample-prompt matching bias}. Specifically, since the prompt serves as a descriptive annotation of the sample class, the ensemble weights for prompt-specific predictions should theoretically be generated based on the prompt-relevant semantics from input samples. However, the image sample contains superfluous semantics (such as background) that are extraneous to the prompt. Therefore, directly utilizing all information from input samples for ensemble weights generation can lead to suboptimal performance, as confirmed by the empirical exploration results in Fig. \ref{fig:motivation2}. To remedy this deficiency, we dexterously extract the prompt-relevant semantics from input samples by introducing the information theory-based regularization constraints. These semantics are then used to compute ensemble weights, which we refer to as debiased ensemble weights because they are derived exclusively from prompt-relevant information. This prevents irrelevant signals (e.g., background color) from influencing the aggregation process, ultimately improving performance.

 

Overall, we propose Adaptive-Debiased Ensemble Multi-Prompt Learning, abbreviated as \textbf{AmPLe}, to jointly mitigate the impact of model-prompt matching bias and sample-prompt matching bias on downstream tasks. AmPLe consists of two key ingredients: (1) hybrid model-prompt ensemble learning module, which leverages the principles of ensemble learning to aggregate the diverse prompt-specific predictions across different VLMs; (2) adaptive-debiased weight generation module, which adaptively calculates debiased ensemble weights by the prompt-relevant information from input samples.
To validate the theoretical soundness of AmPLe, we provide solid mathematical derivations from a causal perspective. 
Benefiting from the plug-and-play nature, the proposed AmPLe is model-agnostic and can be applied on top of a broad spectrum of off-the-shelf VLMs for improving generalization performance. The evaluation results on three representative tasks, namely generalization to novel classes, new target datasets, and unseen domain shifts, fully verify the effectiveness of AmPLe. Our key contributions can be summarized as follows:
\begin{itemize}
    \item We reveal the presence of model-prompt matching bias and sample-prompt matching bias in multi-prompt learning, which negatively affect downstream task inference when using VLMs.
    \item To jointly mitigate model-prompt matching bias and sample-prompt matching bias, we propose a simple yet effective method, namely AmPLe, which incorporates the hybrid model-prompt ensemble learning and adaptive-debiased weight generation modules.
    \item We provide the analysis from a causal perspective, demonstrating the theoretical validity of the proposed method and ensuring its feasibility and reliability.
    \item On three representative tasks, i.e., generalization to novel classes, new target datasets, and unseen domain shifts, the proposed method consistently improves the model’s generalization ability.
\end{itemize}

This paper is organized as follows. Section \ref{related_work} reviews recent advances in VLMs, multi-prompt learning, and ensemble learning. Section \ref{preliminary} introduces the necessary background on applying prompt tuning to downstream classification tasks with CLIP. Section \ref{method} details the design and implementation of our proposed AmPLe. Section \ref{theoretical_analysis} supports the effectiveness of AmPLe with theoretical derivation from a causal perspective. Section \ref{experiment} presents the experimental setup, results, and analysis. Finally, Section \ref{conclusion} summarizes the main contributions of this work and discusses its limitations and future directions.

\section{Related Work} \label{related_work}
In this section, we provide a review of the relevant literature on vision-language models, multi-prompt learning, and ensemble learning.

\subsection{Vision-Language Models} 
Joint learning of vision and language representations processes the two modalities through modality-specific encoders, effectively capturing the distinct characteristics of vision and language. 
By using customized matching constraints~\citep{DBLP:conf/nips/SocherGMN13,DBLP:conf/cvpr/Gomez-BigordaPR17} for further optimization, this approach has demonstrated impressive performance across various downstream tasks. 
Recently, a series of studies~\citep{DBLP:conf/icml/RadfordKHRGASAM21,DBLP:conf/icml/JiaYXCPPLSLD21,DBLP:conf/iclr/LiLZCOSYY22}, notably CLIP~\citep{DBLP:conf/icml/RadfordKHRGASAM21}, train VLMs on large amounts of multi-modal data collected from the web. Specifically, CLIP uses approximately 400 million image-text pairs to train modality-specific encoders in a contrastive manner with the InfoNCE loss~\citep{DBLP:journals/corr/abs-1807-03748}, achieving remarkable zero-shot performance. However, despite the pre-trained VLM acquiring strong generalizable representations, effectively adapting to specific downstream tasks remains an inherent challenge, especially in scenarios with limited training data. To address this challenge, methods~\citep{DBLP:journals/ijcv/ZhouYLL22, DBLP:conf/nips/DuSS24, long2024mutual, DBLP:journals/ijcv/XuZSCLCW25} that combine prompt learning with VLMs have gained widespread attention, demonstrating significant potential for adaptation to a wide range of downstream tasks.

\subsection{Multi-Prompt Learning} 
Prompt learning~\citep{DBLP:conf/nips/BrownMRSKDNSSAA20} was initially introduced in the field of natural language processing and gained prominence with large-scale pre-trained models. It has been demonstrated as an effective method for adapting models to specific downstream tasks by using textual prompts, often without modifying the original model parameters. In recent years, a series of studies~\citep{DBLP:journals/ijcv/ZhouYLL22,DBLP:conf/cvpr/ZhouYL022,DBLP:conf/cvpr/KhattakR0KK23} have shown high accuracy on specific tasks by using learnable prompts on pre-trained VLMs. These methods often rely on a single prompt as the textual prompt, which has limitations for multi-class tasks. As a result, multi-prompt learning, which constructs multiple context templates for prompt learning, has begun to be widely studied. ~\citep{DBLP:conf/cvpr/LuLZL022,derakhshani2022variational} model learnable textual prompts from a distributional perspective, emphasizing the diversity of different textual prompts. ~\citep{DBLP:conf/iclr/0002YSLR023} applies optimal transport~\citep{peyre2019computational} to match vision and text modalities for training multiple comprehensive prompts. ~\citep{DBLP:conf/iccv/ChenHGL023} introduces an energy-based multi-prompt learning method to generate multiple prompt embeddings. 
However, these researches primarily focus on utilizing the multi-prompt learning within a single VLM to achieve superior performance. In this work, we explore the differences in the guiding potential of multiple prompts for different VLMs.

\subsection{Ensemble Learning} 
Ensemble learning is a method that combines the predictions of multiple base learners to improve the overall model performance. Classic ensemble methods, such as Bagging ~\citep{DBLP:journals/ml/Breiman96b,buja2000smoothing}, Boosting~\citep{DBLP:conf/icml/FreundS96,friedman2001greedy,DBLP:conf/icml/CortesMS14}, and Stacking~\citep{DBLP:journals/nn/Wolpert92,DBLP:conf/icassp/DengYP12}, combine multiple weak classifiers using different strategies to enhance accuracy and robustness. Recently, the principles from ensemble learning have been applied across various fields~\citep{DBLP:conf/nips/ChenLSYLKD023,DBLP:journals/ijcv/FangTHY24,DBLP:conf/icml/LuB0XW24,DBLP:journals/ijcv/MuhammadLBO24}, leading to promising advancements. In the task of vision-language inference, ~\citep{DBLP:conf/nips/ChenLSYLKD023} leverages large language models to coordinate multiple VLMs for visual reasoning. Similarly, ~\citep{DBLP:conf/icml/LuB0XW24} combines the outputs of multiple VLMs to achieve superior performance on downstream tasks. In this work, we apply the principles of ensemble learning to aggregate diverse prompt-specific predictions across different VLMs, maximizing the complementary guiding advantages of multiple prompts for different VLMs.

\section{Preliminary} \label{preliminary}
Before introducing AmPLe, we first provide the necessary background on applying prompt tuning to downstream classification tasks with CLIP.
\subsection{Contrastive Language-Image Pretraining (CLIP)} 
CLIP is a foundation VLM pre-trained on approximately 400 million image-text pairs. It consists of a visual branch with a visual encoder \( V(\cdot) \) and a textual branch with a textual encoder \( T(\cdot) \), designed to enable the model to better understand and bridge the semantic gap between visual content and textual descriptions. By aligning the image and text modalities within a unified embedding space using contrastive learning loss, CLIP learns more general visual features, thereby improving its generalization ability. Specifically, let \( C \) denote the number of classes in an image classification task. By appending a specific template, such as ``a photo of a'', in front of each class \( c \in \{1...C\} \), we obtain a set of text inputs \( \{t_c\}_{c=1}^C \). For an image \( x_{\textit{test}} \), its visual feature vector \( z_{\textit{img}} \) is obtained through the visual encoder \( V(\cdot) \), i.e., \( z_{\textit{img}} = V(x_{\textit{test}}) \). Similarly, for the set of text inputs, their corresponding text feature vectors \( \{z_{\textit{text}}^{c}\}_{c=1}^C \) are obtained through the textual encoder \( T(\cdot) \), where \( z_{\textit{text}}^{c} = T(t_c) \). 
Thus, the image classification problem is reformulated as measuring the cosine similarity between the visual feature vector \( z_{\textit{img}} \) and the textual feature vector \( z_{\textit{text}}^{c} \). Consequently, the prediction probability of classifying the image \( x_{\textit{test}} \) as \( c \) is expressed as:
\begin{equation} \label{eq:clip}
  P(c|x_{\textit{test}}) = \frac{\exp(\textit{cos}(z_{\textit{img}}, z_{\textit{text}}^{c}) / \tau)}{\sum_{c'=1}^{C} \exp(\textit{cos}(z_{\textit{img}}, z_{\textit{text}}^{c'}) / \tau)},
\end{equation}
where \( \textit{cos}(\cdot) \) denotes the cosine similarity, and \( \tau \) is the temperature.

\begin{figure*}
    \centering
    \includegraphics[width=0.98\textwidth]{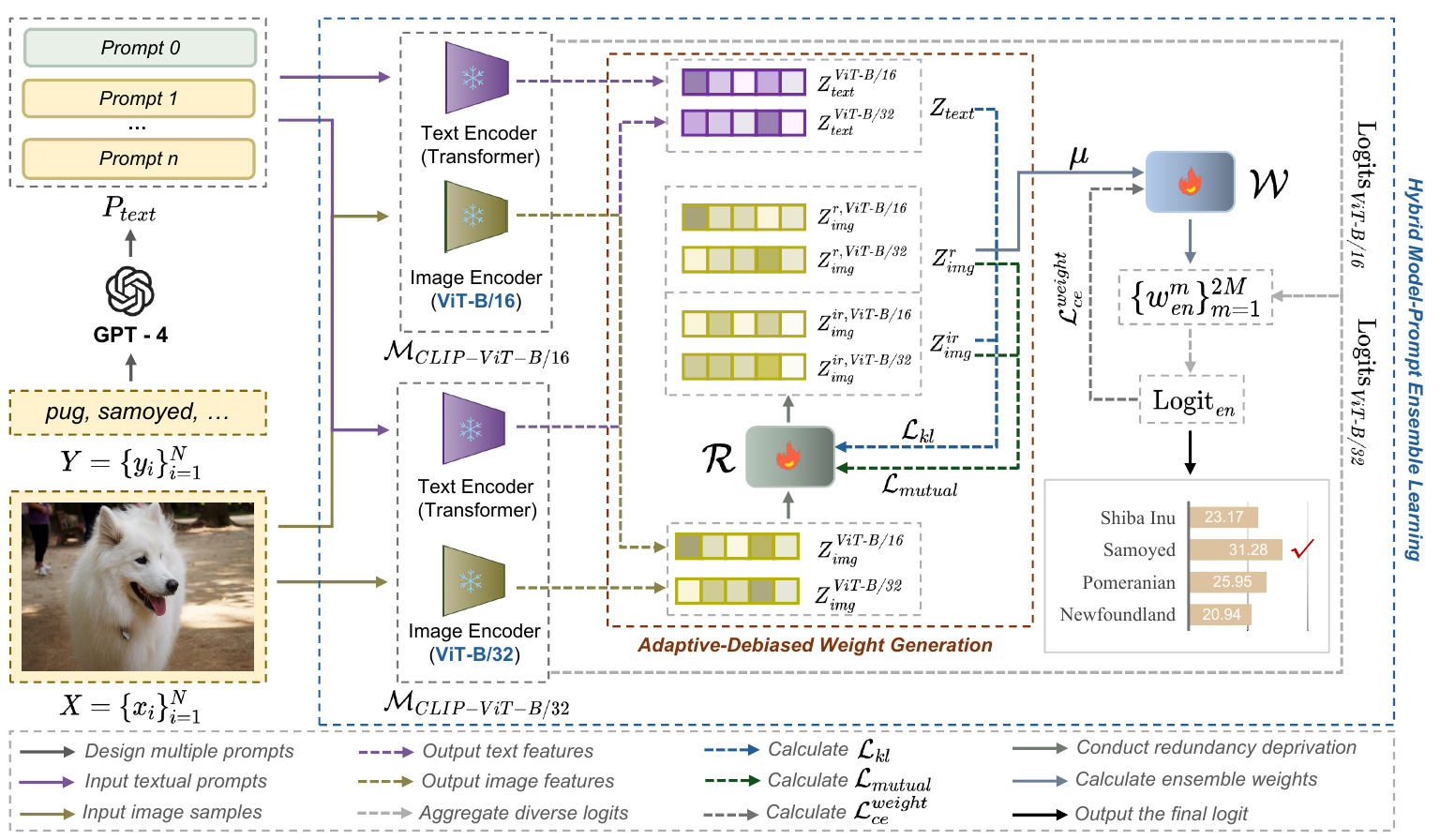}
    \caption{The overall framework of AmPLe. The dashed blue box represents the hybrid model-prompt ensemble learning module, which aggregates diverse predictions from multiple prompts and models to mitigate model-prompt matching bias. The dashed red box represents the adaptive-debiased weight generation module, which extracts prompt-relevant semantics from images to compute debiased ensemble weights, thereby mitigating sample-prompt matching bias. During training, the weight generator \( \mathcal{W} \) and the redundancy deprivation network \( \mathcal{R} \) are optimized using \(\mathcal{L}_{ce}^{weight}\), \(\mathcal{L}_{kl}\), and \(\mathcal{L}_{mutual}\).}
    \label{fig:framework}
\end{figure*}


\subsection{Prompt tuning for CLIP} 
Since CLIP uses manually designed discrete templates, its generalization ability in downstream tasks is potentially limited. In recent years, many studies~\citep{DBLP:journals/ijcv/ZhouYLL22, DBLP:conf/cvpr/ZhouYL022, DBLP:conf/cvpr/KhattakR0KK23, DBLP:conf/iccv/KhattakWNK0K23} have focused on exploring the use of learnable continuous tokens to replace manually designed prompts, and optimizing these learnable tokens with a small amount of labeled data while keeping the parameters of the text encoder and image encoder fixed. Specifically, given a training dataset \( D_{\textit{train}} = \{(x_i, y_i)\}_{i=1}^{N} \), where \( x_i \) represents the \( i \)-th image, \( y_i \in \{1, \dots, C\} \) is the corresponding class label of the image, and \( N \) is the size of the training dataset. We define the learnable prompt as \( {t_c}' = [p_1, p_2, \dots, p_L, e_c] \), where \( p_l \in \{p_l\}_{l=1}^{L} \) represents learnable tokens, \( L \) is the number of context tokens, and \( e_c \) is the word embedding corresponding to class label \( c \). By inputting the learnable prompt into the text encoder \( T(\cdot) \), we obtain a set of learnable text feature vectors \( \{{z_{text}^{c}}’\}_{c=1}^{C} \), where \( {z_{text}^{c}}’ = T({t_c}') \). Therefore, the learnable tokens are optimized across training data by minimizing the following cross-entropy loss function:  
\begin{equation} \label{eq:pt_clip}
    \mathcal{L}_{ce} = -\frac{1}{N}\sum_{i=1}^{N} y_i \log p(y_i | x_i),
\end{equation}
where $p(y_i|x_i) = \frac{\exp(\textit{cos}(z_{\textit{img}}^{i}, {z_{\textit{text}}^{y_i}}’ ) / \tau)}{\sum_{c=1}^{C} \exp(\textit{cos}(z_{\textit{img}}^{i}, {z_{\textit{text}}^{c}}’ ) / \tau)}$, $z_{\textit{img}}^{i} = V(x_i)$, and ${z_{\textit{text}}^{y_i}}’ = T([p_1, p_2, \dots, p_L, e_{y_i}])$.

\section{Methodology} \label{method}
In this section, we first provide an overview of AmPLe in Section \ref{overview}. Then, we introduce the Hybrid Model-Prompt Ensemble Learning module in Section \ref{hmpe}, followed by the Adaptive-Debiased Weight Generation module in Section \ref{adwg}. Finally, we present the objective loss function in Section \ref{final_loss}.

\subsection{Overview} \label{overview}
In this work, we propose the adaptive-debiased ensemble multi-prompt learning (AmPLe) method, which combines multiple prompts and different VLMs to jointly mitigate the impact of model-prompt matching bias and sample-prompt matching bias on downstream tasks. Specifically, we first use GPT-4~\citep{DBLP:journals/corr/abs-2303-08774} to design multiple prompts incorporating different class-relevant visual descriptions from various perspectives of the given class. These prompts are then applied to VLMs trained with CLIP-ViT-B/16 and CLIP-ViT-B/32, and the obtained diverse prompt-specific predictions across different VLMs are aggregated using an ensemble learning approach to maximize the complementary guiding advantages of multiple prompts for different VLMs, thus effectively reducing model-prompt matching bias. For the calculation of ensemble weights, we learn prompt-relevant information from input samples by leveraging the guidance of the information theory-based analysis, enabling the generation of debiased ensemble weights to mitigate sample-prompt matching bias. The overall framework of AmPLe, shown in Figure \ref{fig:framework}, consists of two key ingredients: the hybrid model-prompt ensemble learning module and the adaptive-debiased weight generation module, which are detailed in section \ref{hmpe} and section \ref{adwg}, respectively.

\begin{figure}
    \centering
    \includegraphics[width=0.48\textwidth]{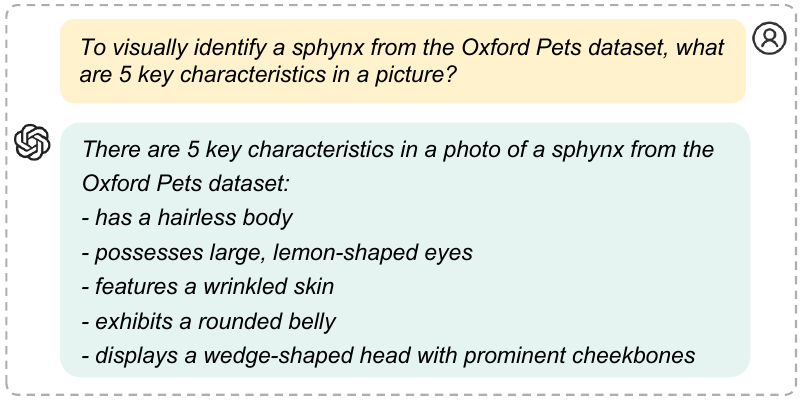}
    \caption{An example of prompting GPT-4 with the query template to generate visual descriptions for the \textit{sphynx} class in the Oxford Pets dataset.}
    \label{fig:generate_prompts}
\end{figure}

\subsection{Hybrid Model-Prompt Ensemble Learning}\label{hmpe} 
In image classification tasks, prompt tuning typically involves introducing learnable contexts before the class name. Recent studies~\citep{DBLP:conf/iclr/MenonV23,DBLP:conf/iccv/RothKKVSA23,DBLP:conf/cvpr/TianZYZ24} have demonstrated that incorporating visual attributes describing the class into prompts can significantly enhance models' generalization ability. As shown in Fig. \ref{fig:generate_prompts}, we leverage GPT-4 to automatically generate multiple visual descriptions of the sphynx class. To design prompts using these visual descriptions, we prepend each visual description with the prefix ``A photo of a sphynx, which'', thereby constructing multiple prompts that capture the characteristics of the sphynx class from various perspectives. As illustrated in Fig. \ref{fig:multi_prompts}, we refer to these prompts, which incorporate different class-relevant descriptors, as domain-relevant semantic prompts. Combining these with the general prompt, i.e., ``A photo of a sphynx.'', we obtain the multiple prompts of the sphynx class. In this work, we design \(M\) prompts that consist of one general prompt and \( M-1 \) domain-relevant semantic prompts for each class \( c \in \{1...C\} \), thus the textual prompts input is formalized as:\begin{equation} \label{eq:text_features}
P_{\textit{text}} = \{\{\textit{prompt}_{c}^{m}\}_{c=1}^{C}\}_{m=1}^{M},
\end{equation}
where \(\textit{prompt}_{c}^{m}\) denotes the \(m\)-th prompt corresponding to the class \(c\). The number of domain-relevant semantic prompts is guided by the empirical analysis presented in Fig. \ref{fig:ablation_n_prompts} of Section \ref{experiment}.

From our empirical exploration results shown in Fig.~\ref{fig:motivation1}, we observed that the identical prompt could exhibit distinct semantic information with respect to different VLMs, thus leading to diverse prompt-specific predictions. We attribute this to model-prompt matching bias and propose a hybrid model-prompt ensemble learning approach to reduce its impact. As highlighted in the dashed blue box of Fig.~\ref{fig:framework}, this module aggregates predictions from multiple prompts across different VLMs. Specifically, to maximize the complementary guiding advantages of multiple prompts for different VLMs, we apply the multiple prompts that incorporate different class-relevant visual descriptions of the given class to different VLMs trained with CLIP-ViT-B/16 and CLIP-ViT-B/32, resulting in diverse prompt-specific prediction logits, image features, and textual features:
\begin{equation} \label{eq:logits_imagefeatures}
    \begin{aligned}
        \text{Logits}_{\textit{ViT-B/16}}, Z_\textit{img}^{\textit{ViT-B/16}}, Z_\textit{text}^{\textit{ViT-B/16}} &= \mathcal{M}_{\textit{CLIP-ViT-B/16}}(P_\textit{text},X), \\
        \text{Logits}_{\textit{ViT-B/32}}, Z_\textit{img}^{\textit{ViT-B/32}}, Z_\textit{text}^{\textit{ViT-B/32}} &= \mathcal{M}_{\textit{CLIP-ViT-B/32}}(P_\textit{text},X),
    \end{aligned}
\end{equation}
where \( X = \{x_i\}_{i=1}^{N} \) denotes the training image input, \( \mathcal{M}_{\textit{CLIP-ViT-B/16}} \) and \( \mathcal{M}_{\textit{CLIP-ViT-B/32}} \) respectively represent the VLMs trained with CLIP-ViT-B/16 and CLIP-ViT-B/32. \( \text{Logits}_{\textit{ViT-B/16}} = \{\text{Logit}_{\textit{ViT-B/16}}^m\}_{m=1}^{M} \) and \( \text{Logits}_{\textit{ViT-B/32}} = \{\text{Logit}_{\textit{ViT-B/32}}^m\}_{m=1}^{M} \) represent the prediction logits obtained from applying the \( M \) prompts to the two VLMs. For simplicity, we use \( \{\text{Logit}_m\}_{m=1}^{2M} \) to represent the \( 2M \) prediction logits. \(Z_\textit{img}^{\textit{ViT-B/16}}\) and \(Z_\textit{img}^{\textit{ViT-B/32}}\) denote the image features, and \(Z_\textit{text}^{\textit{ViT-B/16}}\) and \(Z_\textit{text}^{\textit{ViT-B/32}}\) denote the corresponding text features from the two VLMs.

In our hybrid model-prompt ensemble learning, we employ a weight generator \( \mathcal{W} \) to dynamically generate the ensemble weights:
\begin{equation} \label{eq:logits_weight}
    \{w_{\textit{en}}^m\}_{m=1}^{2M} = \mathcal{W}(\mu),
\end{equation}
where \( \mu \) represents the prompt-relevant information from input samples, which will be detailed in section \ref{adwg}. \(\mathcal{W}\) is designed as a two-layer fully connected network that maps the input to a lower-dimensional hidden representation followed by a ReLU activation, and subsequently projects this hidden representation to produce the weight outputs with a Sigmoid activation applied to bound the generated weights.

The ensemble weights are then used to aggregate \( 2M \) prediction logits, yielding the ensemble prediction logit:
\begin{equation} \label{eq:logits_en}
    \text{Logit}_{\textit{en}} = \sum_{m=1}^{2M} w_{\textit{en}}^m \cdot \text{Logit}_m,
\end{equation}
where \( \text{Logit}_m \in \{\text{Logit}_m\}_{m=1}^{2M} \) represents the \( m \)-th logit from the \( 2M \) prediction logits obtained by applying \( M \) prompts to \( \mathcal{M}_{\textit{CLIP-ViT-B/16}} \) and \( \mathcal{M}_{\textit{CLIP-ViT-B/32}} \), and \( w_{\textit{en}}^m \in \{w_{\textit{en}}^m\}_{m=1}^{2M}\) represents the ensemble weight corresponding to the \( m \)-th logit. 

According to Equation (\ref{eq:pt_clip}), we use the cross-entropy loss to train the weight generator \( \mathcal{W} \). By taking \( \text{Logit}_{\textit{en}} \) and true labels \( Y= \{y_i\}_{i=1}^{N} \) of training image as the input, we compute the cross-entropy loss as follows:
\begin{equation} \label{eq:L_ce_weight}
    \mathcal{L}_{ce}^{weight}=-\frac{1}{N}\sum_{i=1}^{N}y_{i}log(\frac{\exp(\text{Logit}_{en}^{i,y_{i}})}{ {\textstyle \sum_{c=1}^{C}\exp(\text{Logit}_{en}^{i,c})} } ),
\end{equation}
where \( \text{Logit}_{en}^{i,c} \) represents the logit for the \( i \)-th image sample in class \( c \), and \( \text{Logit}_{en}^{i,y_{i}} \) represents the logit for the \( i \)-th image sample in its true class \( y_i \). 

By employing the hybrid model-prompt ensemble learning, we effectively leverage the complementary strengths of multiple prompts and different VLMs, reducing model-prompt matching bias.

\subsection{Adaptive-Debiased Weight Generation}\label{adwg}
For the calculation of ensemble weights, ~\citep{DBLP:conf/icml/LuB0XW24} use the features of input samples to dynamically generate the ensemble weights, which overlooks the existence of sample-prompt matching bias. This bias originates from the prompt-irrelevant semantics encapsulated in input samples, hindering the precision of generated ensemble weights. To tackle this issue, we propose an adaptive-debiased weight generation method, as shown in the dashed red box of Fig.~\ref{fig:framework}. This method first learns prompt-relevant information from input samples by leveraging the guidance of the information theory-based analysis~\citep{ash2012information}, and then uses this information as the input to the weight generator \( \mathcal{W} \) to adaptively compute debiased ensemble weights, thereby reducing sample-prompt matching bias. 

Specifically, we train a redundancy deprivation network \( \mathcal{R} \) to eliminate prompt-irrelevant information from the image feature \( Z_\textit{img} = \{Z_\textit{img}^\textit{ViT-B/16}, Z_\textit{img}^\textit{ViT-B/32}\}\), obtaining the prompt-relevant features \( Z_{\textit{img}}^{r} \) and the prompt-irrelevant features \( Z_{\textit{img}}^{ir} \):
\begin{equation} \label{eq:deprivation}
    Z_{\textit{img}}^{r}, Z_{\textit{img}}^{ir} = \mathcal{R}(Z_\textit{img}),
\end{equation}
where \( Z_{\textit{img}}^{r} = \{z_{\textit{img}}^{r,i}\}_{i=1}^{N} \), \( z_{\textit{img}}^{r,i} = \{z_{\textit{img}}^{r,i, \textit{ViT-B/16}}, z_{\textit{img}}^{r,i, \textit{ViT-B/32}}\} \), \( Z_{\textit{img}}^{ir} = \{z_{\textit{img}}^{ir,i}\}_{i=1}^{N} \), and \( z_{\textit{img}}^{ir,i} = \{z_{\textit{img}}^{ir,i, \textit{ViT-B/16}}, z_{\textit{img}}^{ir,i, \textit{ViT-B/32}}\} \). \( \mathcal{R} \) is implemented as two linear layers with a ReLU activation, motivated by the Universal Approximation Theorem~\citep{DBLP:journals/mcss/Cybenko89, DBLP:journals/nn/Hornik91} and the principle of Occam’s razor~\citep{DBLP:journals/datamine/Domingos99}, which together justify that the design is sufficiently expressive while remaining lightweight.

We assume that prompt relevance is not tied to specific class identities, but instead reflects whether an image feature dimension encodes semantics aligned with the textual prompts. To ensure that prompt-relevant information is only contained in the prompt-relevant image features and not in the prompt-irrelevant image features, we apply regularization constraints to both \( Z_{\textit{img}}^{r} \) and \( Z_{\textit{img}}^{ir} \). Firstly, we minimize the conditional mutual information between the prompt-relevant features \( Z_{\textit{img}}^{r} \) and the prompt-irrelevant features \( Z_{\textit{img}}^{ir} \) given the true labels \( Y \), which is formalized as:
\begin{equation} \label{eq:mul}
\mathcal{L}_{mutual}=I(Z_{\textit{img}}^{r};Z_{\textit{img}}^{ir}|Y),
\end{equation}
where \( I(\cdot) \) represents the Shannon mutual information~\citep{DBLP:journals/bstj/Shannon48}. 

The regularization term \( \mathcal{L}_{mutual} \) effectively reduces the information overlap between the prompt-relevant features \( Z_{\textit{img}}^{r} \) and the prompt-irrelevant features \( Z_{\textit{img}}^{ir} \), thereby ensuring that the information related to prompts exists exclusively in one of the two features. Inspired by ~\citep{DBLP:conf/nips/JiangV22,DBLP:conf/icml/0076M000TH24}, we estimate the conditional mutual information as follows:
\begin{equation} \label{eq:conditional_mul}
    \begin{aligned}
    & I(Z_{\textit{img}}^{r};Z_{\textit{img}}^{ir} \mid Y) \\
    & = \left\| \frac{1}{N} \sum_{i=1}^N z_{\textit{img}}^{r,i} \left( z_{\textit{img}}^{ir,i} - \frac{1}{\left| \# j:y_j=y_i \right|} \sum_{j:y_j=y_i} z_{\textit{img}}^{ir,j}\right) \right\|_1,
    \end{aligned}
\end{equation}
where \(y_i\) corresponds to the class labels covered by the entire training dataset, \(y_j\) refers to the class labels involved in the batch data, and \(\left| \# j:y_j=y_i \right|\) means counting all the numbers that satisfy the condition \(y_j = y_i\).

To enforce the information related to prompts only exists in the prompt-relevant features \( Z_{\textit{img}}^{r}=\{Z_\textit{img}^{r,\textit{ViT-B/16}}, Z_\textit{img}^{r,\textit{ViT-B/32}}\} \), we make the prediction distribution, calculated by combining the prompt-irrelevant features \( Z_{\textit{img}}^{ir}=\{Z_\textit{img}^{ir,\textit{ViT-B/16}}, Z_\textit{img}^{ir,\textit{ViT-B/32}}\} \) with the corresponding textual features \( Z_\textit{text}=\{Z_\textit{text}^{\textit{ViT-B/16}}, Z_\textit{text}^{\textit{ViT-B/32}}\} \), to be close to a uniform distribution. This is formalized as:
\begin{equation} \label{eq:kl}
    \mathcal{L}_{kl}=\sum_{i=1}^{N}D_{KL}\!\left(p_{ir}(y_{i}|x_{i}) \,\middle\|\, p_{uni}\right),
\end{equation}
where \(p_{ir}(y_{i}|x_{i})=\frac{\exp\!\big(\cos(z_{img}^{ir,(\cdot),i}, {z_{text}^{y_{i},(\cdot)}}')/\tau \big)}{\sum_{c=1}^{C}\exp\!\big(\cos(z_{img}^{ir,(\cdot),i}, {z_{text}^{c,(\cdot)}}')/\tau \big)}\), \(p_{uni}=\tfrac{1}{C}\) represents a uniform distribution, and \( C \) is the number of classes. \(D_{KL}\) represents the Kullback-Leibler divergence~\citep{DBLP:conf/iccv/GoldbergerGG03}. \(Z_\textit{text}^{(\cdot)}\) always corresponds to the same VLM that produces \(Z_\textit{img}^{ir,(\cdot)}\), i.e., \(Z_\textit{text}^{\textit{ViT-B/16}}\) is used with \(Z_\textit{img}^{ir,\textit{ViT-B/16}}\), and \(Z_\textit{text}^{\textit{ViT-B/32}}\) is used with \(Z_\textit{img}^{ir,\textit{ViT-B/32}}\).


By jointly minimizing both \( L_{kl} \) and \( L_{mutual} \) during the training process, we effectively eliminate the prompt-irrelevant information from the image sample features. Accordingly, we use prompt-relevant features \( Z_{\textit{img}}^{r} \) to construct the input \(\mu\) of the weight generator \(\mathcal{W} \) for ensemble weights generation, that is:
\begin{equation} \label{eq:concat_input}
    \mu = \textit{concat}(\{Z_{\textit{img}}^{r, \textit{ViT-B/16}}, Z_{\textit{img}}^{r, \textit{ViT-B/32}}\}),
\end{equation}
where \( Z_{\textit{img}}^{r, \textit{ViT-B/16}} \) and \( Z_{\textit{img}}^{r, \textit{ViT-B/32}} \) represent the prompt-relevant features extracted from the image features corresponding to CLIP-ViT-B/16 and CLIP-ViT-B/32, respectively. The function \( \textit{concat}(\cdot) \) denotes the concatenation of \( Z_{\textit{img}}^{r, \textit{ViT-B/16}} \) and \( Z_{\textit{img}}^{r, \textit{ViT-B/32}} \) to meet input requirements of the weight generator \( \mathcal{W} \).

The prompt-relevant information $\mu$ enables the weight generator \( \mathcal{W}\) to focus solely on the image feature related to prompts for ensemble weights generation, thereby achieving the adaptive-debiased weight generation and effectively mitigating sample-prompt matching bias. As the empirical exploration results depicted in Fig. \ref{fig:motivation2}, this design enables the model to concentrate on the target object while largely suppressing attention to irrelevant regions, thereby ensuring performance improvements in practical outcomes.

\begin{algorithm}[!t]
\DontPrintSemicolon
  \SetAlgoLined
	\vskip 0.in
	\begin{algorithmic}
		\STATE {\bfseries Input:} The annotated image datasets \((\mathcal{X}, \mathcal{Y})\), multiple prompts \(P_{\textit{text}} = \{\{\textit{prompt}_{c}^{m}\}_{c=1}^{C}\}_{m=1}^{M}\), hyperparameters $\alpha$ and $\beta$, the minibatch size \(N\), and the total training epoch number \(T\). \\
		\STATE {\bf Initialize:} The learnable weight generator \( \mathcal{W} \), the learnable redundancy deprivation network \( \mathcal{R} \), and the fixed VLMs \( \mathcal{M}_{\textit{CLIP-ViT-B/16}} \) and \( \mathcal{M}_{\textit{CLIP-ViT-B/32}} \).\\
		\For{$t \in [0, T-1]$}{
            \STATE \textbf{Step1.} Iteratively sample a minibatch \(\{(x_i, y_i)\}_{i=1}^{N}\) from \((\mathcal{X}, \mathcal{Y})\), yielding \( X = \{x_i\}_{i=1}^{N} \) and \( Y = \{y_i\}_{i=1}^{N} \);
            \STATE \textbf{Step2.} Obtain diverse prompt-specific prediction logits $\{\text{Logit}_m\}_{m=1}^{2M}$, image features \( Z_\textit{img} \), and textual features \( Z_\textit{text} \) using $\mathcal{M}_{\textit{CLIP-ViT-B/16}}$ and $\mathcal{M}_{\textit{CLIP-ViT-B/32}}$ with \( X \) and \(P_{\textit{text}}\) according to Equation (\ref{eq:logits_imagefeatures});
            \STATE \textbf{Step3.} Use \(\mathcal{R}\) with regularization terms \(\mathcal{L}_{mutual}\) and \(\mathcal{L}_{kl}\) to eliminate prompt-irrelevant information from \( Z_\textit{img} \), according to Equations (\ref{eq:deprivation}), (\ref{eq:mul}), (\ref{eq:conditional_mul}), and (\ref{eq:kl}), yielding prompt-relevant features \( Z_{\textit{img}}^{r} \);
            \STATE \textbf{Step4.} Construct the input \(\mu\) of \(\mathcal{W}\) using \( Z_{\textit{img}}^{r} \) to generate the ensemble weights \(\{w_{\textit{en}}^m\}_{m=1}^{2M}\), and aggregate $\{\text{Logit}_m\}_{m=1}^{2M}$ with \(\{w_{\textit{en}}^m\}_{m=1}^{2M}\) to obtain the ensemble prediction logit \(\text{Logit}_{\textit{en}}\) as described by Equations (\ref{eq:concat_input}), (\ref{eq:logits_weight}), and (\ref{eq:logits_en});
            \STATE \textbf{Step5.} Calculate the cross-entropy loss \(\mathcal{L}_{ce}^{weight}\) using \(\text{Logit}_{\textit{en}}\) and \( Y\), and combine \(\mathcal{L}_{mutual}\) and \(\mathcal{L}_{kl}\) with hyperparameters $\alpha$ and $\beta$ to obtain the final objective loss \(\mathcal{L}\), as described by Equations (\ref{eq:L_ce_weight}) and (\ref{eq:L_final});
            \STATE \textbf{Step6.} Minimize \(\mathcal{L}\) to train both \( \mathcal{R} \) and \( \mathcal{W} \).
            }
	\end{algorithmic}
	\vskip 0in
	\caption{\textbf{:} The training pipeline of AmPLe.}
	\label{alg:ample_train}
\end{algorithm}

\subsection{Final Objective Loss} \label{final_loss}
Combining with Equations (\ref{eq:L_ce_weight}), (\ref{eq:mul}) and (\ref{eq:kl}), we obtain the final objective loss:
\begin{equation} \label{eq:L_final}
    \mathcal{L}=\mathcal{L}_{ce}^{weight}+\alpha \mathcal{L}_{kl}+\beta \mathcal{L}_{mutual},
\end{equation}
where \( \alpha \) and \( \beta \) are hyperparameters to balance the influence of the term \( \mathcal{L}_{kl} \) and the term \( \mathcal{L}_{mutual} \). The training pipeline is detailed by Algorithm \ref{alg:ample_train}.

\section{Theoretical Analysis}\label{theoretical_analysis}

In this section, we demonstrate the effectiveness of the proposed AmPLe with sufficient theoretical support from a causal \citep{pearl2009causality,pearl2016causal} perspective.
Briefly, we attribute the effectiveness of AmPLe to the capture of causal effect between vision-language features and the label, while other VLMs fit the correlation merely. In formal derivation, we adopt the mathematical convention of using uppercase letters to denote random variables and lowercase letters to represent their specific realizations or values.

\begin{figure}
    \centering
    \includegraphics[width=0.3\textwidth]{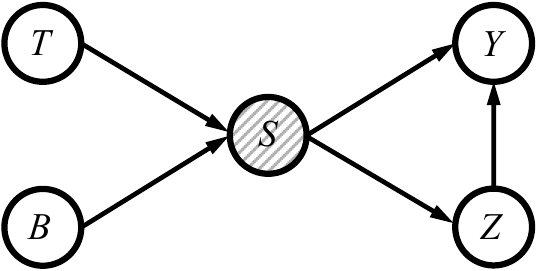}
    \caption{The SCM designed to analyze the inherent mechanism of VLMs.}
    \label{fig:scm}
\end{figure}

To profoundly comprehend the inherent mechanism of utilizing multiple prompts to facilitate the adaptation of VLMs to downstream tasks, from the representation learning perspective, we build the structural causal model (SCM) \citep{pearl2016causal} illustrated in Fig. \ref{fig:scm}, which holds for the following reasons:
\begin{itemize}
    \item[(1)] $T \longrightarrow S \longleftarrow B$: $T$, $B$, and $S$ denote the designed textual prompt, the backbone (VLM), and prompt semantics, respectively. According to empirical results in Fig. \ref{fig:motivation1}, the same prompt can exhibit distinct semantics across different VLMs, thus the prompt semantics are jointly determined by both the designed prompt and VLM. Furthermore, prompt semantics are subjective and abstract concepts, making them inherently unobservable.
    \item[(2)] $S \longrightarrow Y$ and $S \longrightarrow Z \longrightarrow Y$: $Y$ and $Z$ represent the predicted label and vision-language features, respectively. Prompt semantics influence the predicted label through two pathways. First, since the prompt serves as a description of the input sample's class, its semantics directly impact the predicted label $Y$, i.e., $S \longrightarrow Y$ holds. Secondly, different prompt semantics guide the model to focus on different regions of the input sample, leading to the extraction of distinct vision-language features, which are utilized to calculate the predicted $Y$. Consequently, the causal path $S \longrightarrow Z \longrightarrow Y$ exists.
\end{itemize}

Building upon the established SCM, we provide the derivation of causal effect between vision-language features and the label (denoted by $P(y|do(z))$) in the following. 

The joint distribution corresponding to the SCM in Fig. \ref{fig:scm} can be decomposed as:
\begin{equation}
P(t,b,s,y,z) = P(t) P(b) P(s|t,b) P(z|s) P(y|s,z).
\end{equation}
According to the Equation (3.10) in \citep{pearl2009causality}, the $do$-calculus removes the term $P(z|s)$ and induces the post-intervention distribution:
\begin{equation}
P(t,b,s,y|do(z)) = P(t) P(b) P(s|t,b) P(y|s,z).
\end{equation}
Summing over $t,b,s$ gives:
\begin{equation} \label{eq:do-calculus derivation}
\begin{aligned}
   & P(y|do(z))= \sum_t \sum_b \sum_s P(t) P(b) P(s|t,b) P(y|s,z) \\
   & = \sum_t \sum_b \sum_s P(t) P(b) P(s|t,b,z) P(y|s,z,t,b) \\
   & = \sum_t \sum_b P(t) P(b) P(y|z,t,b).
\end{aligned}
\end{equation}
The second subequation in Equation (\ref{eq:do-calculus derivation}) holds due to $S$ is independent of $Z$ given $T$ and $B$, and $Y$ is independent of $T$ and $B$ given $S$ and $Z$. The third subequation in Equation (\ref{eq:do-calculus derivation}) holds due to the law of total probability. 

Drawing upon the example of Simpson’s paradox in \citep{pearl2016causal}, just exploring correlations can yield erroneous results in the presence of the confounder, whereas investigating causal effect will not. In the designed SCM, the variable $S$ functions as a confounder, thus the causal effect between $z$ and $y$ is not equivalent to their correlation, i.e., $p(y|z) \neq p(y|do(z))$. According to Equation (\ref{eq:do-calculus derivation}), capturing the causal effect between $z$ and $y$ requires synthesizing predictions from multiple prompts across different models, which aligns with the design of the hybrid model–prompt ensemble learning module.



\section{Experiment}\label{experiment}
In this section, we conduct extensive experiments to evaluate the effectiveness of the proposed AmPLe. We first introduce the experimental setup (Section \ref{experimental_setup}), including datasets, baselines, and implementation details. Then, we present the main results (Section \ref{main_results}) on various generalization scenarios, and conduct ablation experiments (Section \ref{ablation}) to analyze the impact of key components and multiple prompts. Finally, we provide further analysis (Section \ref{further_analysis}) of significance test, case study, hyperparameters research, and visual comparison.

\subsection{Experimental Setup}\label{experimental_setup}
\textbf{Datasets.} In this work, we build upon previous studies~\citep{DBLP:journals/ijcv/ZhouYLL22,DBLP:conf/cvpr/ZhouYL022,DBLP:conf/cvpr/KhattakR0KK23,DBLP:conf/iccv/KhattakWNK0K23,DBLP:conf/cvpr/YangZWX24} and use 15 image classification datasets to evaluate the effectiveness of the proposed method on three representative tasks, i.e., generalization to novel classes, new target datasets and unseen domain shifts. To assess our method's performance in base-to-novel and cross-dataset generalization, we use 11 of the image classification datasets, which cover a variety of classification tasks. These include five fine-grained datasets, i.e., OxfordPets~\citep{DBLP:conf/cvpr/ParkhiVZJ12}, StanfordCars~\citep{DBLP:conf/iccvw/Krause0DF13}, Flowers102~\citep{DBLP:conf/icvgip/NilsbackZ08}, Food101~\citep{DBLP:conf/eccv/BossardGG14}, and Aircraft~\citep{DBLP:journals/corr/MajiRKBV13}; two general object datasets, i.e., ImageNet~\citep{DBLP:conf/cvpr/DengDSLL009} and Caltech101~\citep{DBLP:conf/cvpr/LiFP04}; a texture classification dataset, i.e., DTD~\citep{DBLP:conf/cvpr/CimpoiMKMV14}; a satellite image dataset, i.e., EuroSAT~\citep{DBLP:journals/staeors/HelberBDB19}; a scene recognition dataset, i.e., SUN397~\citep{DBLP:conf/cvpr/XiaoHEOT10}; and an action recognition dataset, i.e., UCF101~\citep{DBLP:journals/corr/abs-1212-0402}. For evaluating domain generalization, we use ImageNet as the source dataset and select four variant datasets of ImageNet, i.e., ImageNetV2~\citep{DBLP:conf/icml/RechtRSS19}, ImageNet-Sketch~\citep{DBLP:conf/nips/WangGLX19}, ImageNet-A~\citep{DBLP:conf/cvpr/HendrycksZBSS21}, and ImageNet-R~\citep{DBLP:conf/iccv/HendrycksBMKWDD21}, as target datasets. Table~\ref{tab:dataset-overview} summarizes the statistics of these datasets.

\begin{table}[t]
\centering
\small
\setlength{\tabcolsep}{2.7mm}
\caption{Statistics of datasets used in experiments.}
\begin{tabular}{lcccc}
\toprule
Dataset & \#Classes & \#Train  & \#Val & \#Test \\
\midrule
ImageNet & 1,000 & 1.28M & N/A & 50,000\\
ImageNet-A & 200 & N/A & N/A & 7,500\\
ImageNet-R & 200 & N/A & N/A & 30,000\\
ImageNet-Sketch & 1,000 & N/A & N/A & 50,889\\
ImageNet-V2 & 1,000 & N/A & N/A & 10,000\\
\midrule
Caltech101 & 100 & 4128 & 1649 & 2,465\\
OxfordPets & 37 & 2944 & 736 & 3,669\\
StanfordCars & 196 & 6509 & 1635 & 8,041\\
Flowers102 & 102 & 4093 & 1633 & 2,463\\
Food101 & 101 & 50500 & 20200 & 30,300\\
Aircraft & 100 & 3334 & 3333 & 3,333\\
SUN397 & 397 & 15880 & 3970 & 19,850 \\
DTD & 47 & 2820 & 1128 & 1,692\\
EuroSAT & 10 & 13500 & 5400 & 8,100\\
UCF101 & 101 & 7639 & 1898 & 3,783 \\
\bottomrule
\end{tabular}
\label{tab:dataset-overview}
\end{table}

\textbf{Baselines.} Our primary baselines include MaPLe~\citep{DBLP:conf/cvpr/KhattakR0KK23}, MMA~\citep{DBLP:conf/cvpr/YangZWX24}, and MMRL~\citep{DBLP:conf/cvpr/GuoG25}. MaPLe learns prompts in both vision and language branches for tighter alignment. MMA introduces a multi-modal adapter to improve the alignment between the textual and visual representations. MMRL~\citep{DBLP:conf/cvpr/GuoG25} establishes a shared, learnable, and modality-agnostic representation space, projecting space tokens to text and image representations to facilitate more effective multi-modal interactions. Benefiting from the plug-and-play nature, we apply the proposed AmPLe to the three baselines to evaluate its effectiveness. In addition, we compare our method with CLIP~\citep{DBLP:conf/icml/RadfordKHRGASAM21}, CoOp~\citep{DBLP:journals/ijcv/ZhouYLL22}, CoCoOp~\citep{DBLP:conf/cvpr/ZhouYL022}, IPO~\citep{DBLP:conf/nips/DuSS24}, DiMPLe~\citep{DBLP:journals/corr/abs-2506-21237}, and AttriPrompt~\citep{zhan2025attriprompt}. CLIP demonstrates strong zero-shot performance, CoOp illustrates the effectiveness of vanilla prompt tuning, and CoCoOp highlights the benefit of image-conditioned prompt tuning. IPO~\citep{DBLP:conf/nips/DuSS24} leverages large language models for interpretable prompt generation, DiMPLe~\citep{DBLP:journals/corr/abs-2506-21237} disentangles invariant and spurious features for OOD robustness, and AttriPrompt~\citep{zhan2025attriprompt} composes prompts dynamically from visual attributes. These comparisons allow us to thoroughly assess our method's performance across different settings, providing a comprehensive evaluation of its effectiveness in prompt learning scenarios.

\textbf{Implementation Details.} We follow the experimental settings of ~\citep{DBLP:conf/icml/LuB0XW24}. Specifically, for the data loader settings, we set the batch size of the training set to 128 and the batch size of the test set to 100. Additionally, we set 8 worker threads to speed up data loading. Regarding the optimizer, we use Stochastic Gradient Descent (SGD) with a learning rate of \( 2e-2 \), and the maximum number of training epochs is set to 5. To adjust the learning rate during training, a cosine learning rate scheduler is employed to gradually decrease the learning rate during the training process. Furthermore, we perform a constant learning rate warm-up of 1 epoch at the beginning of training to stabilize training in the early stages, and the learning rate during the warm-up phase is set to \( 1e-5 \). In experiments, all models are trained on a single NVIDIA 4090 GPU, and the reported results are averaged over three random runs. The experiments are conducted in a few-shot setting, where each class is trained with only 16 labeled samples. For the base-to-novel generalization setup, the model is trained exclusively on the base classes in a 16-shot setting, while evaluation is conducted on both base and novel classes. The evaluation metrics include the classification accuracy for both base and novel classes, as well as their Harmonic Mean (HM). For cross-dataset generalization, the model is trained with 16 labeled samples per class on all 1,000 classes of ImageNet and subsequently evaluated on other datasets to assess generalization performance. Similarly, for domain generalization, the model is trained on the original ImageNet dataset and evaluated on four of its variants, each incorporating different types of domain shifts. The evaluation metric for both cross-dataset and domain generalization experiments is classification accuracy.


\begin{table*}[]
  \caption{Comparison with baseline methods on the base-to-novel generalization setting. $\dag$ represents our reproduction result of the corresponding official code.}
  \label{tab:AMpLe_base2new}
  \centering
\adjustbox{max width=\textwidth}{
    \begin{tabular}{ll|lll|lll|lll|lll}
    \midrule
    \multicolumn{2}{c|}{\multirow{2}[2]{*}{}} & \multicolumn{3}{c|}{\textbf{Average}} & \multicolumn{3}{c|}{\textbf{ImageNet}} & \multicolumn{3}{c|}{\textbf{Caltech101}} & \multicolumn{3}{c}{\textbf{OxfordPets}} \\
    \multicolumn{2}{c|}{} & \multicolumn{1}{c}{\textbf{Base}} & \multicolumn{1}{c}{\textbf{New}} & \multicolumn{1}{c|}{\textbf{HM}} & \multicolumn{1}{c}{\textbf{Base}} & \multicolumn{1}{c}{\textbf{New}} & \multicolumn{1}{c|}{\textbf{HM}} & \multicolumn{1}{c}{\textbf{Base}} & \multicolumn{1}{c}{\textbf{New}} & \multicolumn{1}{c|}{\textbf{HM}} & \multicolumn{1}{c}{\textbf{Base}} & \multicolumn{1}{c}{\textbf{New}} & \multicolumn{1}{c}{\textbf{HM}} \\
    \midrule[0.8pt]
    \multicolumn{2}{l|}{CLIP~\cite{DBLP:conf/icml/RadfordKHRGASAM21}} & 69.34  & 74.22  & 71.70  & 72.43  & 68.14  & 70.22  & 96.84  & 94.00  & 95.40  & 91.17  & 97.26  & 94.12  \\
    \multicolumn{2}{l|}{CoOp~\cite{DBLP:journals/ijcv/ZhouYLL22}} & 82.69  & 63.22  & 71.66  & 76.47  & 67.88  & 71.92  & 98.00  & 89.81  & 93.73  & 93.67  & 95.29  & 94.47  \\
    \multicolumn{2}{l|}{CoCoOp~\cite{DBLP:conf/cvpr/ZhouYL022}} & 80.47  & 71.69  & 75.83  & 75.98  & 70.43  & 73.10  & 97.96  & 93.81  & 95.84  & 95.20  & 97.69  & 96.43  \\
    \multicolumn{2}{l|}{MaPLe~\cite{DBLP:conf/cvpr/KhattakR0KK23}} & 82.28 & 75.14 & 78.55  & 76.66 & 70.54 & 73.47 & 97.74 & 94.36 & 96.02 & 95.43 & 97.76 & 96.58  \\
    \multicolumn{2}{l|}{MaPLe$^{\dag}$~\cite{DBLP:conf/cvpr/KhattakR0KK23}} & 82.23 & 74.82 & 78.35  & 77.00 & 70.70 & 73.72 & 98.10 & 94.87 & 96.46 & 95.53 & 97.73 & 96.62  \\
    \multicolumn{2}{l|}{IPO~\cite{DBLP:conf/nips/DuSS24}} & 79.92 & 80.51 & 80.21  & 77.83 & 72.45 & 75.04 & 97.32 & 95.23 & 96.26 & 95.21 & 98.23 & 96.70  \\
    \multicolumn{2}{l|}{MMA~\cite{DBLP:conf/cvpr/YangZWX24}} & 83.20 & 76.94 & 79.95 & 77.31 & 71.00 & 74.02 & 98.40 & 94.00 & 96.15 & 95.40 & 98.07 & 96.72 \\
    \multicolumn{2}{l|}{MMA$^{\dag}$~\cite{DBLP:conf/cvpr/YangZWX24}} & 83.16 & 76.78 & 79.85 & 77.43 & 71.03 & 74.10  & 98.27 & 94.03 & 96.10 & 95.47 & 98.10 & 96.77 \\
    \multicolumn{2}{l|}{DiMPLe~\cite{DBLP:journals/corr/abs-2506-21237}} & 76.09 & 73.35 & 74.70 & 71.87 & 62.67 & 66.96 & 97.43 & 93.53 & 95.44 & 91.57 & 97.73 & 94.55 \\
    \multicolumn{2}{l|}{AttriPrompt~\cite{zhan2025attriprompt}} & 84.88 & 77.63 & 81.09 & 77.63 & 71.00 & 74.17 & 98.50 & 95.53 & 96.99 & 96.13 & 98.07 & 97.09 \\
    \multicolumn{2}{l|}{MMRL~\cite{DBLP:conf/cvpr/GuoG25}} & 85.68 & 77.16 & 81.20 & 77.90 & 71.30 & 74.45  & 98.97 & 94.50 & 96.68 & 95.90 & 97.60 & 96.74 \\
    \multicolumn{2}{l|}{MMRL$^{\dag}$~\cite{DBLP:conf/cvpr/GuoG25}} & 85.68 & 76.78 & 80.99 & 77.87 & 71.13 & 74.35  & 98.97 & 94.50 & 96.68 & 95.83 & 97.97 & 96.89 \\
    \midrule
    \rowcolor{mygray}
    \multicolumn{2}{l|}{MaPLe  + AmPLe} & 83.85 & 77.00 & 80.28 & 77.90 & 71.33 & 74.47 & 98.33 & 96.03 & 97.17 & 95.47 & 96.87 & 96.16 \\
    \multicolumn{2}{l|}{\textit{Performance Gains}} & \textcolor{red}{+1.62} & \textcolor{red}{+2.18} & \textcolor{red}{+1.93} & \textcolor{red}{+0.90} & \textcolor{red}{+0.63} & \textcolor{red}{+0.75} & \textcolor{red}{+0.23} & \textcolor{red}{+1.16} & \textcolor{red}{+0.71} & \textcolor{blue}{-0.06} & \textcolor{blue}{-0.86} & \textcolor{blue}{-0.46} \\ 
    \rowcolor{mygray}
    \multicolumn{2}{l|}{MMA  + AmPLe} & 84.49 & 77.99 & 81.11 & 78.67 & 72.03 & 75.20 & 98.50 & 95.03 & 96.74  & 95.73 & 97.10 & 96.41 \\
    \multicolumn{2}{l|}{\textit{Performance Gains}} & \textcolor{red}{+1.33} & \textcolor{red}{+1.21} & \textcolor{red}{+1.26} & \textcolor{red}{+1.24} & \textcolor{red}{+1.00} & \textcolor{red}{+1.10} & \textcolor{red}{+0.23} & \textcolor{red}{+1.00} & \textcolor{red}{+0.64} & \textcolor{red}{+0.26} & \textcolor{blue}{-1.00} & \textcolor{blue}{-0.36} \\
    \rowcolor{mygray}
    \multicolumn{2}{l|}{MMRL  + AmPLe} & 86.35 & 79.08 & 82.56 & 78.50 & 72.43 & 75.34 & 99.00 & 94.87 & 96.89  & 95.40 & 97.90 & 96.63 \\
    \multicolumn{2}{l|}{\textit{Performance Gains}} & \textcolor{red}{+0.67} & \textcolor{red}{+2.30} & \textcolor{red}{+1.57} & \textcolor{red}{+0.63} & \textcolor{red}{+1.30} & \textcolor{red}{+0.99} & \textcolor{red}{+0.03} & \textcolor{red}{+0.37} & \textcolor{red}{+0.21} & \textcolor{blue}{-0.43} & \textcolor{blue}{-0.07} & \textcolor{blue}{-0.26} \\
\midrule
    \multicolumn{2}{c|}{\multirow{2}[2]{*}{}} & \multicolumn{3}{c|}{\textbf{StanfordCars}} & \multicolumn{3}{c|}{\textbf{Flowers102}} & \multicolumn{3}{c|}{\textbf{Food101}} & \multicolumn{3}{c}{\textbf{Aircraft}} \\
    \multicolumn{2}{c|}{} & \multicolumn{1}{c}{\textbf{Base}} & \multicolumn{1}{c}{\textbf{New}} & \multicolumn{1}{c|}{\textbf{HM}} & \multicolumn{1}{c}{\textbf{Base}} & \multicolumn{1}{c}{\textbf{New}} & \multicolumn{1}{c|}{\textbf{HM}} & \multicolumn{1}{c}{\textbf{Base}} & \multicolumn{1}{c}{\textbf{New}} & \multicolumn{1}{c|}{\textbf{HM}} & \multicolumn{1}{c}{\textbf{Base}} & \multicolumn{1}{c}{\textbf{New}} & \multicolumn{1}{c}{\textbf{HM}} \\
    \midrule[0.8pt]
    \multicolumn{2}{l|}{CLIP~\cite{DBLP:conf/icml/RadfordKHRGASAM21}} & 63.37  & 74.89  & 68.65  & 72.08  & 77.80  & 74.83  & 90.10  & 91.22  & 90.66  & 27.19  & 36.29 & 31.09  \\
    \multicolumn{2}{l|}{CoOp~\cite{DBLP:journals/ijcv/ZhouYLL22}} & 78.12  & 60.40  & 68.13  & 97.60  & 59.67  & 74.06  & 88.33  & 82.26  & 85.19  & 40.44  & 22.30  & 28.75  \\
    \multicolumn{2}{l|}{CoCoOp~\cite{DBLP:conf/cvpr/ZhouYL022}} & 70.49  & 73.59  & 72.01  & 94.87  & 71.75  & 81.71  & 90.70 & 91.29  & 90.99  & 33.41  & 23.71  & 27.74  \\
    \multicolumn{2}{l|}{MaPLe~\cite{DBLP:conf/cvpr/KhattakR0KK23}} & 72.94 & 74.00 & 73.47 & 95.92 & 72.46 & 82.56 & 90.71 & 92.05 & 91.38 & 37.44 & 35.61 & 36.50 \\
    \multicolumn{2}{l|}{MaPLe$^{\dag}$~\cite{DBLP:conf/cvpr/KhattakR0KK23}} & 72.87 & 73.80 & 73.33  & 96.27 & 73.23 & 83.19 & 90.77 & 92.17 & 91.46 & 37.67 & 35.73 & 36.67  \\
    \multicolumn{2}{l|}{IPO~\cite{DBLP:conf/nips/DuSS24}} & 73.42 & 75.71 & 74.55  & 96.78 & 78.32 & 86.58 & 90.92 & 93.08 & 91.99 & 41.21 & 41.42 & 41.31  \\
    \multicolumn{2}{l|}{MMA~\cite{DBLP:conf/cvpr/YangZWX24}} & 78.50 & 73.10 & 75.70 & 97.77 & 75.93 & 85.48 & 90.13 & 91.30 & 90.71 & 40.57 & 36.33 & 38.33 \\
    \multicolumn{2}{l|}{MMA$^{\dag}$~\cite{DBLP:conf/cvpr/YangZWX24}} & 78.47 & 72.93 & 75.60 & 97.77 & 75.63 & 85.29  & 90.13 & 91.33 & 90.73 & 41.10 & 35.43 & 38.06 \\
    \multicolumn{2}{l|}{DiMPLe~\cite{DBLP:journals/corr/abs-2506-21237}} & 67.97 & 74.07 & 70.89 & 84.33 & 75.00 & 79.39 & 89.80 & 91.23 & 90.51 & 30.57 & 36.80 & 33.40 \\
    \multicolumn{2}{l|}{AttriPrompt~\cite{zhan2025attriprompt}} & 80.33 & 75.50 & 77.70 & 98.33 & 77.40 & 86.62 & 90.77 & 91.93 & 91.35 & 42.97 & 37.07 & 39.80 \\
    \multicolumn{2}{l|}{MMRL~\cite{DBLP:conf/cvpr/GuoG25}} & 81.30 & 75.07 & 78.06 & 98.97 & 77.27 & 86.78 & 90.57 & 91.50 & 91.03 & 46.30 & 37.03 & 41.15 \\
    \multicolumn{2}{l|}{MMRL$^{\dag}$~\cite{DBLP:conf/cvpr/GuoG25}} & 81.37 & 74.80 & 77.95 & 98.83 & 76.97 & 86.54 & 90.63 & 91.53 & 91.08 & 45.97 & 37.27 & 41.17 \\
    \midrule
    \rowcolor{mygray}
    \multicolumn{2}{l|}{MaPLe  + AmPLe} & 76.63 & 75.53 & 76.08 & 98.27 & 78.43 & 87.24 & 90.77 & 91.87 & 91.31 & 37.63 & 36.03 & 36.82 \\
    \multicolumn{2}{l|}{\textit{Performance Gains}} & \textcolor{red}{+3.76} & \textcolor{red}{+1.73} & \textcolor{red}{+2.75} & \textcolor{red}{+2.00} & \textcolor{red}{+5.20} & \textcolor{red}{+4.05} & \textcolor{blue}{-0.00} & \textcolor{blue}{-0.30} & \textcolor{blue}{-0.15} & \textcolor{blue}{-0.04} & \textcolor{red}{+0.30} & \textcolor{red}{+0.15} \\
    \rowcolor{mygray}
    \multicolumn{2}{l|}{MMA  + AmPLe} & 81.70 & 76.17 & 78.84 & 98.80 & 77.83 & 87.07 & 90.67 & 91.70 & 91.18 & 42.93 & 38.13 & 40.39 \\
    \multicolumn{2}{l|}{\textit{Performance Gains}} & \textcolor{red}{+3.23} & \textcolor{red}{+3.24} & \textcolor{red}{+3.24} & \textcolor{red}{+1.03} & \textcolor{red}{+2.20} & \textcolor{red}{+1.78} & \textcolor{red}{+0.54} & \textcolor{red}{+0.37} & \textcolor{red}{+0.45} & \textcolor{red}{+1.83} & \textcolor{red}{+2.70} & \textcolor{red}{+2.33} \\
    \rowcolor{mygray}
    \multicolumn{2}{l|}{MMRL  + AmPLe} & 83.37 & 78.13 & 80.66 & 99.07 & 78.73 & 87.74 & 90.50 & 91.93 & 91.21 & 48.37 & 40.90 & 44.32 \\
    \multicolumn{2}{l|}{\textit{Performance Gains}} & \textcolor{red}{+2.00} & \textcolor{red}{+3.33} & \textcolor{red}{+2.71} & \textcolor{red}{+0.24} & \textcolor{red}{+1.76} & \textcolor{red}{+1.20} & \textcolor{blue}{-0.13} & \textcolor{red}{+0.40} & \textcolor{red}{+0.13} & \textcolor{red}{+2.40} & \textcolor{red}{+3.63} & \textcolor{red}{+3.15} \\
    \midrule
    \multicolumn{2}{c|}{\multirow{2}[2]{*}{}} & \multicolumn{3}{c|}{\textbf{SUN397}} & \multicolumn{3}{c|}{\textbf{DTD}} & \multicolumn{3}{c|}{\textbf{EuroSAT}} & \multicolumn{3}{c}{\textbf{UCF101}} \\
    \multicolumn{2}{c|}{} & \multicolumn{1}{c}{\textbf{Base}} & \multicolumn{1}{c}{\textbf{New}} & \multicolumn{1}{c|}{\textbf{HM}} & \multicolumn{1}{c}{\textbf{Base}} & \multicolumn{1}{c}{\textbf{New}} & \multicolumn{1}{c|}{\textbf{HM}} & \multicolumn{1}{c}{\textbf{Base}} & \multicolumn{1}{c}{\textbf{New}} & \multicolumn{1}{c|}{\textbf{HM}} & \multicolumn{1}{c}{\textbf{Base}} & \multicolumn{1}{c}{\textbf{New}} & \multicolumn{1}{c}{\textbf{HM}} \\
    \midrule[0.8pt]
    \multicolumn{2}{l|}{CLIP~\cite{DBLP:conf/icml/RadfordKHRGASAM21}} & 69.36  & 75.35  & 72.23  & 53.24  & 59.90  & 56.37  & 56.48  & 64.05  & 60.03  & 70.53  & 77.50  & 73.85  \\
    \multicolumn{2}{l|}{CoOp~\cite{DBLP:journals/ijcv/ZhouYLL22}} & 80.60  & 65.89  & 72.51  & 79.44  & 41.18  & 54.24  & 92.19  & 54.74  & 68.69  & 84.69  & 56.05  & 67.46  \\
    \multicolumn{2}{l|}{CoCoOp~\cite{DBLP:conf/cvpr/ZhouYL022}} & 79.74  & 76.86  & 78.27  & 77.01  & 56.00  & 64.85  & 87.49  & 60.04  & 71.21  & 82.33  & 73.45  & 77.64  \\
    \multicolumn{2}{l|}{MaPLe~\cite{DBLP:conf/cvpr/KhattakR0KK23}} & 80.82 & 78.70 & 79.75 & 80.36 & 59.18 & 68.16 & 94.07 & 73.23 & 82.35 & 83.00 & 78.66 & 80.77 \\
    \multicolumn{2}{l|}{MaPLe$^{\dag}$~\cite{DBLP:conf/cvpr/KhattakR0KK23}} & 80.73 & 78.60 & 79.65  & 79.93 & 54.87 & 65.07 & 92.30 & 74.27 & 82.31 & 83.33 & 77.07 & 80.08  \\
    \multicolumn{2}{l|}{IPO~\cite{DBLP:conf/nips/DuSS24}} & 81.25 & 80.92 & 81.08  & 82.14 & 66.81 & 73.69 & 94.25 & 80.11 & 86.61 & 85.32 & 80.92 & 83.06  \\
    \multicolumn{2}{l|}{MMA~\cite{DBLP:conf/cvpr/YangZWX24}} & 82.27 & 78.57 & 80.38 & 83.20 & 65.63 & 73.38 & 85.46 & 82.34 & 83.87 & 86.23 & 80.03 & 83.01 \\
    \multicolumn{2}{l|}{MMA$^{\dag}$~\cite{DBLP:conf/cvpr/YangZWX24}} & 82.30 & 78.53 & 80.37 & 83.10 & 65.63 & 73.34  & 84.60 & 83.40 & 84.00 & 86.13 & 78.57 & 82.18 \\
    \multicolumn{2}{l|}{DiMPLe~\cite{DBLP:journals/corr/abs-2506-21237}} & 75.43 & 74.27 & 74.85 & 69.80 & 62.30 & 65.84 & 80.10 & 63.93 & 71.11 & 78.10 & 75.33 & 76.69 \\
    \multicolumn{2}{l|}{AttriPrompt~\cite{zhan2025attriprompt}} & 82.77 & 79.50 & 81.10 & 84.87 & 65.03 & 73.64 & 93.87 & 81.27 & 87.12 & 87.50 & 81.63 & 84.46 \\
    \multicolumn{2}{l|}{MMRL~\cite{DBLP:conf/cvpr/GuoG25}} & 83.20 & 79.30 & 81.20 & 85.67 & 65.00 & 73.92 & 95.60 & 80.17 & 87.21 & 88.10 & 80.07 & 83.89 \\
    \multicolumn{2}{l|}{MMRL$^{\dag}$~\cite{DBLP:conf/cvpr/GuoG25}} & 83.13 & 79.23 & 81.13 & 85.77 & 64.23 & 73.45 & 95.77 & 77.40 & 85.61 & 88.37 & 79.50 & 83.70 \\
    \midrule
    \rowcolor{mygray}
    \multicolumn{2}{l|}{MaPLe  + AmPLe} & 82.57 & 79.80 & 81.16 & 81.80 & 62.57 & 70.90 & 96.17 & 79.53 & 87.06 & 86.87 & 79.00 & 82.75 \\
    \multicolumn{2}{l|}{\textit{Performance Gains}} & \textcolor{red}{+1.84} & \textcolor{red}{+1.20} & \textcolor{red}{+1.51} & \textcolor{red}{+1.87} & \textcolor{red}{+7.70} & \textcolor{red}{+5.83} & \textcolor{red}{+3.87} & \textcolor{red}{+5.26} & \textcolor{red}{+4.75} & \textcolor{red}{+3.54} & \textcolor{red}{+1.93} & \textcolor{red}{+2.67} \\
    \rowcolor{mygray}
    \multicolumn{2}{l|}{MMA  + AmPLe} & 83.67 & 80.00 & 81.79 & 84.53 & 67.03 & 74.77 & 86.17 & 84.07 & 85.10 & 88.03 & 78.83 & 83.18 \\
    \multicolumn{2}{l|}{\textit{Performance Gains}} & \textcolor{red}{+1.37} & \textcolor{red}{+1.47} & \textcolor{red}{+1.42} & \textcolor{red}{+1.43} & \textcolor{red}{+1.40} & \textcolor{red}{+1.43} & \textcolor{red}{+1.57} & \textcolor{red}{+0.67} & \textcolor{red}{+1.10} & \textcolor{red}{+1.90} & \textcolor{red}{+0.26} & \textcolor{red}{+1.00} \\
    \rowcolor{mygray}
    \multicolumn{2}{l|}{MMRL  + AmPLe} & 83.97 & 80.67 & 82.29 & 85.50 & 68.90 & 76.31 & 96.37 & 82.27 & 88.76 & 89.83 & 83.10 & 86.33 \\
    \multicolumn{2}{l|}{\textit{Performance Gains}} & \textcolor{red}{+0.84} & \textcolor{red}{+1.44} & \textcolor{red}{+1.16} & \textcolor{blue}{-0.27} & \textcolor{red}{+4.67} & \textcolor{red}{+2.86} & \textcolor{red}{+0.60} & \textcolor{red}{+4.87} & \textcolor{red}{+3.15} & \textcolor{red}{+1.46} & \textcolor{red}{+3.60} & \textcolor{red}{+2.63} \\
    \midrule
    \end{tabular}%
    
  }
\end{table*}

\begin{table*}[]
    \centering
    \caption{Comparison with baseline methods on the cross-dataset generalization setting.}
    \label{tab:AmPLe_xd} 
    \adjustbox{max width=\textwidth}{
    \begin{tabular}{l l lllllllllll}
    \toprule
    & \textbf{Source} & \multicolumn{11}{c}{\textbf{Target}} \\ \cmidrule(lr){2-2} \cmidrule(lr){3-13}
    & \rotatebox{90}{ImageNet} & \rotatebox{90}{Caltech101} & \rotatebox{90}{OxfordPets} & \rotatebox{90}{StanfordCars} & \rotatebox{90}{Flowers102} & \rotatebox{90}{Food101} & \rotatebox{90}{Aircraft} & \rotatebox{90}{SUN397} & \rotatebox{90}{DTD} & \rotatebox{90}{EuroSAT} & \rotatebox{90}{UCF101} & \rotatebox{90}{\emph{Average}} \\
    \midrule
    CLIP \cite{DBLP:conf/icml/RadfordKHRGASAM21} & 66.72 & 92.94 & 89.07 & 65.29 & 71.30 & 86.11 & 24.87 & 62.62 & 44.56 & 47.69 & 66.77 & 65.12 \\
    CoOp \cite{DBLP:journals/ijcv/ZhouYLL22} & 71.51 & 93.70 & 89.14 & 64.51 & 68.71 & 85.30 & 18.47 & 64.15 & 41.92 & 46.39 & 66.55 & 63.88 \\
    CoCoOp \cite{DBLP:conf/cvpr/ZhouYL022} & 71.02 & 94.43 & 90.14 & 65.32 & 71.88 & 86.06 & 22.94 & 67.36 & 45.73 & 45.37 & 68.21 & 65.74 \\
    MaPLe \cite{DBLP:conf/cvpr/KhattakR0KK23} & 70.72 & 93.53 & 90.49 & 65.57 & 72.23 & 86.20 & 24.74 & 67.01 & 46.49 & 48.06 & 68.69 & 66.30  \\
    MaPLe$^{\dag}$ \cite{DBLP:conf/cvpr/KhattakR0KK23} & 70.50 & 93.73 & 89.83 & 65.23 & 70.97 & 86.07 & 23.67 & 67.03 & 45.47 & 45.27 & 68.27 & 65.55  \\
    IPO \cite{DBLP:conf/nips/DuSS24} & 72.15 & 94.34 & 90.96 & 66.10 & 72.75 & 86.75 & 25.14 & 67.97 & 47.01 & 48.56 & 69.23 & 67.36  \\
    MMA \cite{DBLP:conf/cvpr/YangZWX24} & 71.00 & 93.80 & 90.30 & 66.13 & 72.07 & 86.12 & 25.33 & 68.17 & 46.57 & 49.24 & 68.32 & 66.61 \\
    MMA$^{\dag}$ \cite{DBLP:conf/cvpr/YangZWX24} & 73.03 & 93.17 & 90.10 & 64.20 & 69.80 & 84.87 & 23.07 & 67.23 & 45.37 & 36.50 & 68.60 & 64.29 \\
    AttriPrompt~\cite{zhan2025attriprompt} & 72.40 & 94.23 & 90.73 & 65.63 & 71.57 & 86.47 & 24.37 & 68.00 & 48.37 & 53.17 & 69.17 & 67.17 \\
    MMRL \cite{DBLP:conf/cvpr/GuoG25} & 72.03 & 94.67 & 91.43 & 66.10 & 72.77 & 86.40 & 26.30 & 67.57 & 45.90 & 53.10 & 68.27 & 67.25 \\
    MMRL$^{\dag}$ \cite{DBLP:conf/cvpr/GuoG25} & 72.00 & 94.77 & 91.23 & 66.37 & 72.27 & 86.47 & 25.93 & 67.43 & 47.00 & 51.65 & 68.87 & 67.20 \\
    \midrule
    \rowcolor{mygray}
    MaPLe  + AmPLe & 70.87 & 94.77 & 90.53 & 63.63 & 72.20 & 85.40 & 24.20 & 66.97 & 45.83 & 56.93 & 71.03 & 67.15 \\
    \textit{Performance Gains} & \textcolor{red}{+0.37} & \textcolor{red}{+1.04} & \textcolor{red}{+0.70} & \textcolor{blue}{-1.60} & \textcolor{red}{+1.23} & \textcolor{blue}{-0.67} & \textcolor{red}{+0.53} & \textcolor{blue}{-0.06} & \textcolor{red}{+0.36} & \textcolor{red}{+11.66} & \textcolor{red}{+2.76} & \textcolor{red}{+1.60} \\
    \rowcolor{mygray}
    MMA  + AmPLe & 72.70 & 94.90 & 90.53 & 67.07 & 74.13 & 86.27 & 24.47 & 69.30 & 45.40 & 45.20 & 71.57 & 66.88 \\
    \textit{Performance Gains} & \textcolor{blue}{-0.33} & \textcolor{red}{+1.73} & \textcolor{red}{+0.43} & \textcolor{red}{+2.87} & \textcolor{red}{+4.33} & \textcolor{red}{+1.40} & \textcolor{red}{+1.40} & \textcolor{red}{+2.07} & \textcolor{red}{+0.03} & \textcolor{red}{+8.70} & \textcolor{red}{+2.97} & \textcolor{red}{+2.59} \\
    \rowcolor{mygray}
    MMRL  + AmPLe & 73.17 & 95.17 & 90.30 & 69.83 & 73.90 & 86.70 & 26.93 & 69.97 & 51.13 & 53.30 & 71.50 & 68.87 \\
    \textit{Performance Gains} & \textcolor{red}{+1.17} & \textcolor{red}{+0.40} & \textcolor{blue}{-0.93} & \textcolor{red}{+3.46} & \textcolor{red}{+1.63} & \textcolor{red}{+0.23} & \textcolor{red}{+1.00} & \textcolor{red}{+2.54} & \textcolor{red}{+4.13} & \textcolor{red}{+1.65} & \textcolor{red}{+2.63} & \textcolor{red}{+1.67} \\
    \bottomrule
    \end{tabular}
    }
\end{table*}

\begin{table*}[]
\caption{Comparison with baseline methods on the domain generalization setting.}
\label{tab:AMpLe_dg}
\centering
    \setlength{\tabcolsep}{3.3mm}
    \begin{tabular}{l llllll}
    \toprule
    & \textbf{Source} & \multicolumn{5}{c}{\textbf{Target}} \\ \cmidrule(lr){2-2} \cmidrule(lr){3-7}
     & ImageNet & ImageNetV2 & ImageNet-Sketch & ImageNet-A & ImageNet-R & Average \\
    \midrule
    CLIP \cite{DBLP:conf/icml/RadfordKHRGASAM21} &  66.73 & 60.83 & {46.15} & 47.77 & {73.96} & 57.18 \\
    CoOp \cite{DBLP:journals/ijcv/ZhouYLL22} &  71.51 & 64.20 & 47.99  & 49.71 & 75.21 & 59.28  \\
    CoCoOp \cite{DBLP:conf/cvpr/ZhouYL022} & 71.02 & {64.07} & 48.75 & 50.63 & 76.18 & 59.91 \\
    MaPLe \cite{DBLP:conf/cvpr/KhattakR0KK23} & 70.72 & {64.07} & 49.15 & 50.90  & 76.98 & 60.28 \\
    MaPLe$^{\dag}$ \cite{DBLP:conf/cvpr/KhattakR0KK23} & 70.50 & {63.73} & 48.87 & 51.03  & 77.07 & 60.18 \\
    MMA \cite{DBLP:conf/cvpr/YangZWX24} & 71.00 & 64.33 & 49.13 & 51.12 & 77.32 & 60.48 \\
    MMA$^{\dag}$ \cite{DBLP:conf/cvpr/YangZWX24} & 73.03 & 65.77 & 48.70 & 48.10 & 76.00 & 59.64 \\
    DiMPLe \cite{DBLP:journals/corr/abs-2506-21237} & 69.73 & 61.20 & 45.67 & 44.07 & 73.87 & 56.20 \\
    AttriPrompt~\cite{zhan2025attriprompt} & 72.40 & 65.07 & 49.60 & 52.07 & 78.17 & 61.23 \\
    MMRL \cite{DBLP:conf/cvpr/GuoG25} & 72.03 & 64.47 & 49.17 & 51.20 & 77.53 & 60.59 \\
    MMRL$^{\dag}$ \cite{DBLP:conf/cvpr/GuoG25} & 72.00 & 64.70 & 49.33 & 50.90 & 77.47 & 60.60 \\
    \midrule
    \rowcolor{mygray}
    MaPLe  + AmPLe & 70.87 & 64.43 & 50.37 & 48.93  & 77.90 & 60.41 \\
    \textit{Performance Gains} & {\textcolor{red}{+0.37}} & {\textcolor{red}{+0.70}} & {\textcolor{red}{+1.50}} & {\textcolor{blue}{-2.10}} & {\textcolor{red}{+0.83}} & {\textcolor{red}{+0.23}} \\
    \rowcolor{mygray}
    MMA  + AmPLe & 72.70 & 65.73 & 50.57 & 47.77 & 78.07 & 60.53 \\
    \textit{Performance Gains} & \textcolor{blue}{-0.33} & \textcolor{blue}{-0.04} & \textcolor{red}{+1.87} & \textcolor{blue}{-0.33} & \textcolor{red}{+2.07} & \textcolor{red}{+0.89} \\
    \rowcolor{mygray}
    MMRL  + AmPLe & 73.17 & 65.53 & 50.77 & 48.67 & 78.30 & 60.82 \\
    \textit{Performance Gains} & \textcolor{red}{+1.17} & \textcolor{red}{+0.83} & \textcolor{red}{+1.44} & \textcolor{blue}{-2.23} & \textcolor{red}{+0.83} & \textcolor{red}{+0.22} \\
    \bottomrule
    \end{tabular}
\end{table*}

\subsection{Main Results} \label{main_results}
\textbf{Base-to-Novel Generalization.} Table \ref{tab:AMpLe_base2new} presents a performance comparison of integrating AmPLe into three recent works: MaPLe~\citep{DBLP:conf/cvpr/KhattakR0KK23}, MMA~\citep{DBLP:conf/cvpr/YangZWX24}, and MMRL~\citep{DBLP:conf/cvpr/GuoG25}. In terms of average performance, methods integrated with AmPLe consistently perform better across most datasets. Their Harmonic Mean (HM) significantly surpasses that of baseline methods in the majority of datasets, indicating a better balance between base and novel classes. Specifically, MaPLe + AmPLe outperforms the baseline MaPLe in both base and novel class performance across 8 datasets, and its HM exceeds that of MaPLe in 9 datasets. 
MMA + AmPLe outperforms the baseline MMA in base class performance across all 11 datasets, and in novel class performance and HM values, it surpasses MMA in 10 datasets. MMRL+AmPLe achieves the strongest performance overall, improving HM on 10 out of 11 datasets and raising the average HM to 82.56\% (+1.57\% over MMRL). These consistent gains confirm that AmPLe complements different adaptation paradigms, from prompt learning (MaPLe) to adapter-based tuning (MMA) and representation-token learning (MMRL). Nevertheless, AmPLe does not yield uniform gains across all datasets. For example, on OxfordPets, methods integrated with AmPLe slightly underperform their baselines. We attribute this to the fact that OxfordPets contains highly distinctive and homogeneous categories with limited domain shift, where a single strong prompt already suffices. In such cases, AmPLe’s ensemble weighting introduces minor trade-offs without clear benefit. Importantly, these drops are small (\(< 0.5\) in HM), whereas on more challenging datasets like DTD, EuroSAT, and UCF101, MaPLe + AmPLe achieves improvements of 5.83\%, 4.75\%, and 2.67\% in HM, respectively, while MMRL + AmPLe achieves improvements of 2.86\%, 3.15\%, and 2.63\% in HM. These results further validate the effectiveness of the proposed AmPLe.

\textbf{Cross-Dataset Generalization.} The cross-dataset generalization experiment setup trains the model on ImageNet and then evaluates it on 10 target datasets. As shown in Table~\ref{tab:AmPLe_xd}, integrating AmPLe leads to consistent improvements. Specifically, MaPLe+AmPLe surpasses the baseline MaPLe on 7 of the 10 target datasets with an average gain of +1.6\%, while MMA+AmPLe improves over MMA on all 10 targets, achieving an average gain of +2.59\%. MMRL+AmPLe also delivers solid improvements (+1.67\% on average), confirming that AmPLe is compatible with different adaptation strategies. We observe that the gains are most striking under challenging distribution shifts. For example, MaPLe+AmPLe and MMA+AmPLe improve EuroSAT by +11.66\% and +8.70\%, respectively, suggesting that AmPLe is particularly beneficial when the test domain differs substantially from ImageNet. This is because MaPLe and MMA rely on fixed prompt structures or adapters that constrain flexibility, leading to mismatches between learned representations and new domains. On fine-grained datasets such as StanfordCars and Flowers102, MMA+AmPLe yields notable boosts (+2.87\% and +4.33\%), indicating that AmPLe better distinguishes subtle inter-class variations. Overall, these results suggest that AmPLe effectively mitigates challenges posed by domain and dataset variations, thereby facilitating stronger cross-dataset generalization.

\textbf{Domain Generalization.} The domain generalization experiment setup transfers models trained on ImageNet to four challenging variants: ImageNetV2, ImageNet-Sketch, ImageNet-A, and ImageNet-R. As shown in Table~\ref{tab:AMpLe_dg}, MaPLe+AmPLe improves over MaPLe by +0.7\% on ImageNetV2, +1.5\% on ImageNet-Sketch, and +0.83\% on ImageNet-R, while MMA+AmPLe achieves larger gains of +1.87\% on ImageNet-Sketch and +2.07\% on ImageNet-R. MMRL+AmPLe also shows consistent improvements (+1.44\% on ImageNet-Sketch, +0.83\% on ImageNetV2 and ImageNet-R), reaching the highest average accuracy of 60.82\%. Notably, the gains on ImageNet-A are limited, with slight decreases. This is likely because ImageNet-A emphasizes adversarial corner cases, where even adaptive weighting cannot fully compensate for the absence of similar cues during training. Nevertheless, in terms of overall performance, AmPLe consistently raises the average accuracy of MaPLe, MMA, and MMRL by +0.23\%, +0.89\%, and +0.22\%, respectively, demonstrating its effectiveness in handling domain generalization tasks.

\subsection{Ablation Experiments} \label{ablation}

\begin{table}[tb]
    \caption{Ablation study on base-to-novel generalization.}
    \centering
    \label{tab:AmPLe_ablation}
    \setlength{\tabcolsep}{1.0mm}
    \begin{tabular}{c|cc|ccc}
    \toprule
     & HMPE & ADWG & Base & New & HM \\
    \midrule
    \multirow{3}{*}{MaPLe  + AmPLe} & $\checkmark$ & $\checkmark$ & 83.85 & 77.00 & 80.28 \\
    & $\checkmark$ & $\times$ & 83.73 & 76.68 & 80.05  \\
    & $\times$ & $\times$ & 83.03 & 75.82 & 79.26  \\
    \multirow{3}{*}{MMA  + AmPLe} & $\checkmark$ & $\checkmark$ & 84.49 & 77.99 & 81.11   \\
    & $\checkmark$ & $\times$ & 84.49 & 77.84 & 81.03  \\
    & $\times$ & $\times$ & 84.00 & 77.00 & 80.35   \\
    \multirow{3}{*}{MMRL  + AmPLe} & $\checkmark$ & $\checkmark$ & 86.35 & 79.08 & 82.56   \\
    & $\checkmark$ & $\times$ & 86.16 & 78.80 & 82.32  \\
    & $\times$ & $\times$ & 86.19 & 78.31 & 82.06   \\
    \bottomrule
    \end{tabular}
\end{table}

\begin{figure}
    \centering
    \includegraphics[width=0.49\textwidth]{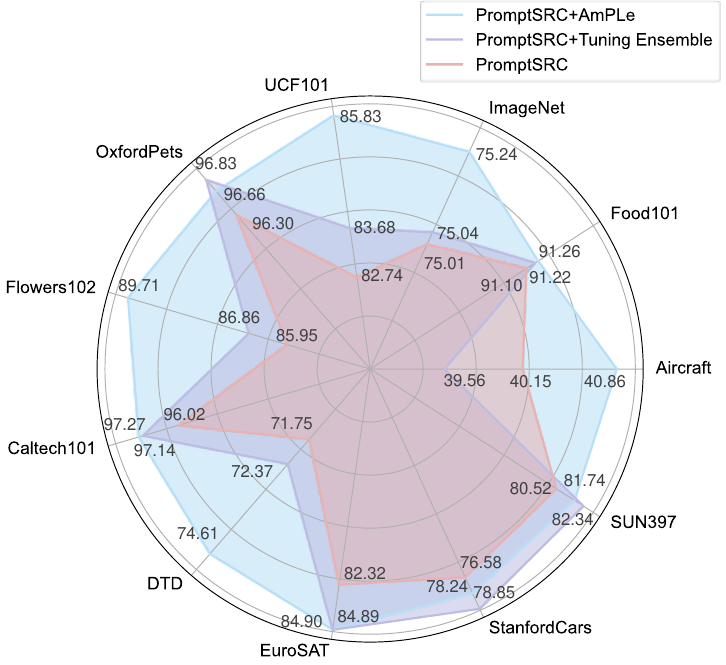}
    \caption{Performance comparison of integrating AmPLe and Tuning Ensemble into the baseline PromptSRC.}
    \label{fig:tuningen_ample}
\end{figure}

\textbf{Components Ablation.} We assess the impact of removing two components, i.e., hybrid model-prompt ensemble learning (HMPE) and adaptive-debiased weight generation (ADWG), on model performance in the base-to-novel generalization setting. As shown in Table \ref{tab:AmPLe_ablation}, the experimental results highlight the contribution of each component. Specifically, when both HMPE and ADWG are included, MaPLe + AmPLe achieves a base class accuracy of 83.85\%, a novel class accuracy of 77.00\%, and a HM value of 80.28\%. Removing ADWG causes a slight drop in performance, with the base class accuracy decreasing to 83.73\%, the novel class accuracy to 76.68\%, and HM to 80.05\%. When both HMPE and ADWG are removed, the performance of MaPLe + AmPLe declines further, with the base class accuracy dropping to 83.03\%, the novel class accuracy to 75.82\%, and HM to 79.26\%, indicating that both HMPE and ADWG contribute positively on the model's performance. Similarly, for MMA + AmPLe, the best performance is achieved when both components are included, further highlighting the importance of HMPE and ADWG in improving the model's overall classification performance. Additionally, we compare AmPLe with the state-of-the-art method Tuning Ensemble~\citep{DBLP:conf/icml/LuB0XW24}, which generates weights based on sample features for weaker VLMs ensemble. Figure \ref{fig:tuningen_ample} presents a performance comparison of integrating these two methods into the baseline PromptSRC~\citep{DBLP:conf/iccv/KhattakWNK0K23}. It can be observed that both methods improve the performance of the baseline PromptSRC, but AmPLe performs more excellently, surpassing Tuning Ensemble on 7 out of the 11 datasets. This further confirms the effectiveness of our proposed hybrid model-prompt ensemble learning module and adaptive-debiased weight generation module.

\begin{figure}
    \centering
    \includegraphics[width=0.48\textwidth]{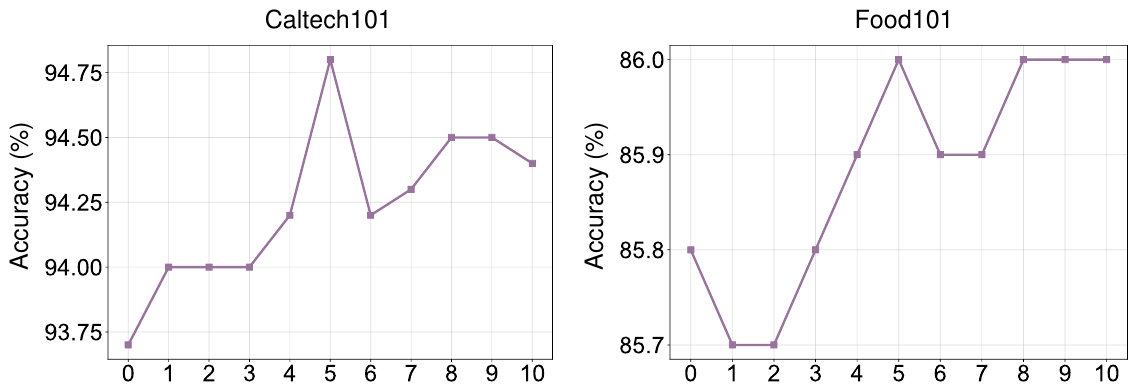}
    \caption{Zero-shot accuracy variations when using different numbers of domain-relevant semantic prompts with CLIP-ViT-B/16 on the Caltech101 and Food101 datasets.}
    \label{fig:ablation_n_prompts}
\end{figure}

\begin{table}[t]
    \caption{Comparison of MMRL+AmPLe with and without shuffle prompts.}
    \centering
    \label{tab:shuffle_ablation}
    \setlength{\tabcolsep}{1.2mm}
    \begin{tabular}{c|ccc|ccc}
    \toprule
    \multirow{2}{*}{} & \multicolumn{3}{c|}{w/o shuffle} & \multicolumn{3}{c}{w/ shuffle} \\
    & Base & New & HM & Base & New & HM \\
    \midrule
    Caltech101      & 99.00 & 94.87 & 96.89 & 98.97 & 94.87 & 96.88 \\
    DTD             & 85.50 & 68.90 & 76.31 & 85.47 & 69.00 & 76.36 \\
    EuroSAT         & 96.37 & 82.27 & 88.76 & 96.40 & 82.30 & 88.79 \\
    Aircraft   & 48.37 & 40.90 & 44.32 & 48.40 & 40.77 & 44.26 \\
    Food101         & 90.50 & 91.93 & 91.21 & 90.50 & 91.90 & 91.19 \\
    ImageNet        & 78.50 & 72.43 & 75.34 & 78.47 & 72.43 & 75.33 \\
    Flowers102  & 99.07 & 78.73 & 87.74 & 99.03 & 78.73 & 87.72 \\
    OxfordPets     & 95.40 & 97.90 & 96.63 & 95.40 & 97.90 & 96.63 \\
    StanfordCars   & 83.37 & 78.13 & 80.66 & 83.40 & 78.13 & 80.68 \\
    SUN397          & 83.97 & 80.67 & 82.29 & 84.00 & 80.67 & 82.30 \\
    UCF101          & 89.83 & 83.10 & 86.33 & 89.83 & 83.20 & 86.39 \\
    \bottomrule
    \end{tabular}
\end{table}

\textbf{Multiple Prompts Ablation.}
We conduct an ablation study using the CLIP-ViT-B/16 to investigate the impact of incorporating multiple domain-relevant semantic prompts on zero-shot classification performance. The results, depicted in Fig. \ref{fig:ablation_n_prompts}, illustrate the accuracy variations when employing different numbers of prompts, ranging from 0 to 10, on the Caltech101 and Food101 datasets. For the Caltech101 dataset, we observe a general upward trend in accuracy as the number of domain-relevant semantic prompts increases, with a notable peak at five prompts. Beyond this point, the accuracy slightly fluctuates but remains relatively stable, suggesting that excessive domain-relevant semantic prompts do not necessarily yield further performance gains. A similar pattern is observed on the Food101 dataset, where accuracy initially improves as more domain-relevant semantic prompts are incorporated. The performance stabilizes when using five or more prompts, indicating a saturation effect. This trend implies that leveraging a moderate number of domain-relevant semantic prompts effectively enhances zero-shot classification performance by enriching the semantic space, while excessive prompts could introduce redundancy or noise. Based on the empirical results, we design multiple prompts for each class, including five domain-relevant semantic prompts and one general prompt, to balance the advantages of leveraging domain-specific knowledge while maintaining generalization, thus ensuring optimal performance across diverse datasets. 

\textbf{Prompt Order Ablation.} 
To further investigate the influence of semantic ordering on the effectiveness of AmPLe, we conduct an ablation study by randomly shuffling the order of the prompts within each class. This perturbation disrupts any potential positional consistency across prompts while preserving their semantic content. As shown in Table~\ref{tab:shuffle_ablation}, the performance of MMRL+AmPLe remains nearly unchanged after shuffling. This result confirms that the proposed method does not rely on maintaining cross-class prompt order alignment. Instead, AmPLe benefits from the overall semantic diversity of prompts, while the adaptive-debiased weight generation module effectively conditions the ensemble weights on prompt-relevant image features. These findings validate the robustness of AmPLe against variations in prompt ordering and highlight that its improvements mainly stem from leveraging diverse semantic descriptors rather than assuming strict positional semantics.

\begin{figure}
    \centering
    \includegraphics[width=0.49\textwidth]{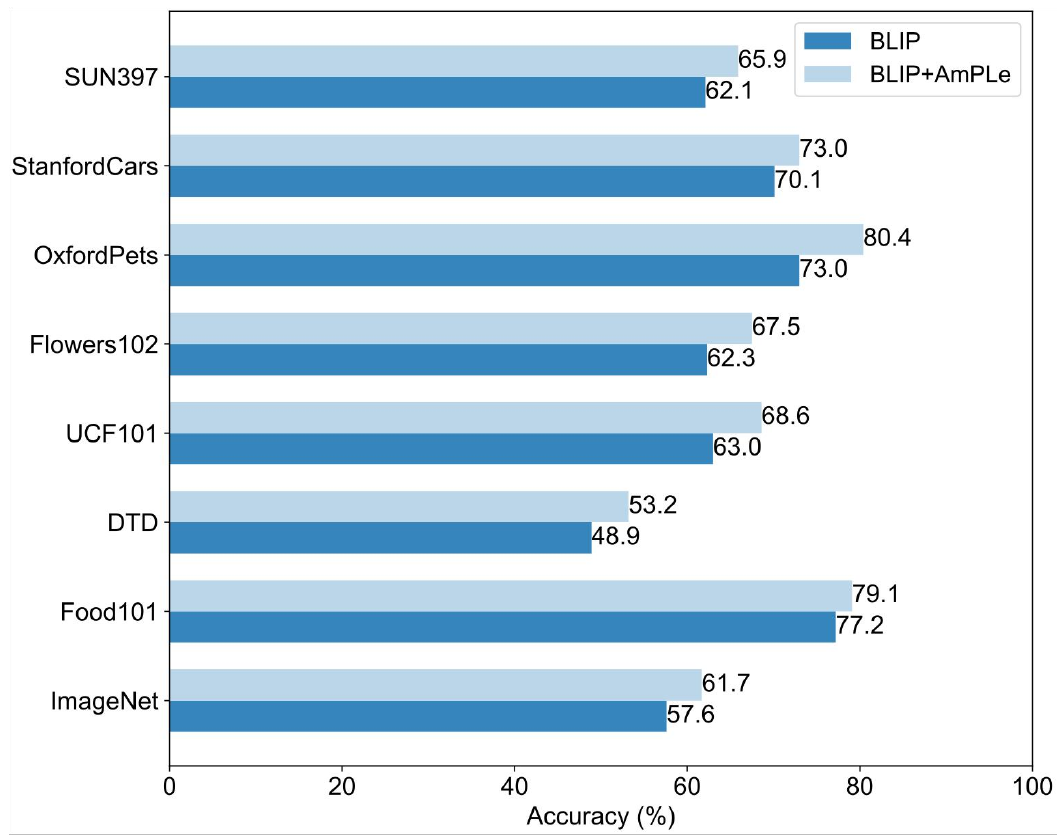}
    \caption{Zero-shot performance comparison between BLIP and BLIP+AmPLe across different datasets.}
    \label{fig:blip_ample}
\end{figure}

\subsection{Further Analysis} 
\textbf{Comparison with BLIP on Zero-Shot Performance} We further evaluate AmPLe on BLIP~\citep{DBLP:conf/icml/0001LXH22}, a representative model that advances cross-modal alignment via multimodal pre-training and image-grounded language modeling. Specifically, we report the zero-shot performance of BLIP and BLIP+AmPLe, as shown in Fig.~\ref{fig:blip_ample}. Incorporating AmPLe yields consistent improvements across datasets, with notable gains on SUN397 (+3.8), DTD (+4.3), and UCF101 (+5.6). These results indicate that AmPLe is not limited to CLIP-style contrastive models but also enhances zero-shot cross-modal alignment in more advanced architectures. This demonstrates the generality and novelty of our approach: AmPLe functions as a complementary mechanism that seamlessly strengthens both CLIP-like and BLIP-style models, thereby delivering more robust and transferable performance across tasks and domains.

\begin{table}[tb]
    \caption{The \textit{p}-value for the Student's \textit{t}-test across three task settings. BNG, CDG, and DG are abbreviations for Base-to-Novel Generalization, Cross-Dataset Generalization, and Domain Generalization, respectively.}
    \centering
    \label{tab:t_test}
    \setlength{\tabcolsep}{4.6mm}
    \begin{tabular}{c|ccc}
    \toprule
    &  & \textit{p}-value & \\
    \cmidrule(lr){2-4}
    & BNG & CDG & DG \\
    \midrule
    MaPLe + AmPLe & \multirow{2}{*}{0.046} & \multirow{2}{*}{0.023} & \multirow{2}{*}{0.037} \\
    \textit{versus MaPLe} &  &  & \\
    MMA + AmPLe & \multirow{2}{*}{0.004} & \multirow{2}{*}{0.003} & \multirow{2}{*}{0.030} \\
    \textit{versus MMA} & & & \\
    MMRL + AmPLe & \multirow{2}{*}{0.021} & \multirow{2}{*}{0.014} & \multirow{2}{*}{0.018} \\
    \textit{versus MMRL} & & & \\
    \bottomrule
    \end{tabular}
\end{table}

\begin{table}[tb]
    \caption{Complexity analysis of AmPLe compared with baselines on the base-to-novel generalization setting. ``Params'' denotes the number of trainable parameters (in millions), and ``FPS'' denotes the inference throughput (frames per second), which is calculated as the number of test samples divided by the total inference time.}
    \centering
    \label{tab:AmPLe_complexity}
    \setlength{\tabcolsep}{1.9mm}
    \begin{tabular}{c|ccc|cc}
    \toprule
     & Base & New & HM & Params & FPS  \\
    \midrule
    {MaPLe} & 82.23 & 74.82 & 78.35 & 3.55M & 420 \\
    {MaPLe  + AmPLe} & 83.85 & 77.00 & 80.28 & 6.73M & 233 \\
    {MMA} & 83.16 & 76.78 & 79.85 & 0.67M & 381   \\
    {MMA  + AmPLe} & 84.49 & 77.99 & 81.11 & 3.85M & 221   \\
    {MMRL} & 85.68 & 76.78 & 80.98 & 4.99M & 131 \\
    {MMRL  + AmPLe} & 86.35 & 79.08 & 82.55 & 8.17M & 105 \\
    \bottomrule
    \end{tabular}
\end{table}

\textbf{Significance Test.} To ensure that the observed performance improvements are not due to randomness, we conduct a Student’s t-test~\citep{mendenhall2020introduction} between the baseline models and their counterparts integrated with AmPLe. A p-value below 0.05 indicates that the performance improvements are statistically significant compared to the baseline models. As shown in Table~\ref{tab:t_test}, all p-values across the three task settings, i.e., Base-to-Novel Generalization (BNG), Cross-Dataset Generalization (CDG), and Domain Generalization (DG), are consistently below this threshold. Specifically, for the comparison between MaPLe + AmPLe and MaPLe, the p-values are 0.046, 0.023, and 0.037, respectively. For MMA + AmPLe versus MMA, the values are 0.004, 0.003, and 0.030, while for MMRL + AmPLe versus MMRL, the values are 0.021, 0.014, and 0.018. These results consistently confirm that the integration of AmPLe yields statistically significant performance gains across different task settings, further highlighting its robustness and demonstrating that the improvements are not incidental but systematically enhance the models’ generalization ability.

\begin{figure}
    \centering
    \includegraphics[width=0.49\textwidth]{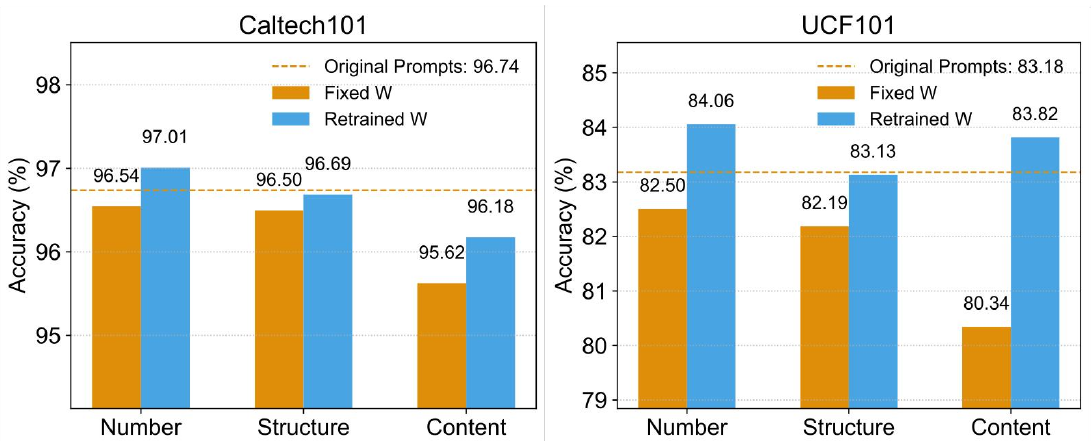}
    \caption{Adaptability of MMA+AmPLe under different prompt sets. ``Fixed \( \mathcal{W} \)'' denotes directly reusing the weight generator trained with the original prompt set, while ``Retrained \( \mathcal{W} \)'' indicates retraining it on the new prompt set.}
    \label{fig:prompts_variant}
\end{figure}

\textbf{Computational Complexity Analysis.} 
For a more comprehensive evaluation of the trade-off between accuracy gains and computational overhead, we perform complexity analyses under the base-to-novel generalization setting, with results summarized in Table~\ref{tab:AmPLe_complexity}. Integrating AmPLe into recent baselines (MaPLe, MMA, and MMRL) consistently improves the Harmonic Mean (HM) by approximately 1.2–1.9\% on average. In terms of efficiency, AmPLe introduces a moderate increase in the number of trainable parameters compared to the baseline models, while maintaining the same order of magnitude, and it reduces inference throughput (e.g., from 381 FPS to 221 FPS for MMA, and from 131 FPS to 105 FPS for MMRL). Despite the increases in model size and latency, the additional overhead remains acceptable given the performance improvements achieved. Moreover, the statistically significant gains across three generalization tasks (\(p < 0.05\) in Table \ref{tab:t_test}) further justify the added complexity.

\begin{figure*}
    \centering
    \includegraphics[width=0.99\textwidth]{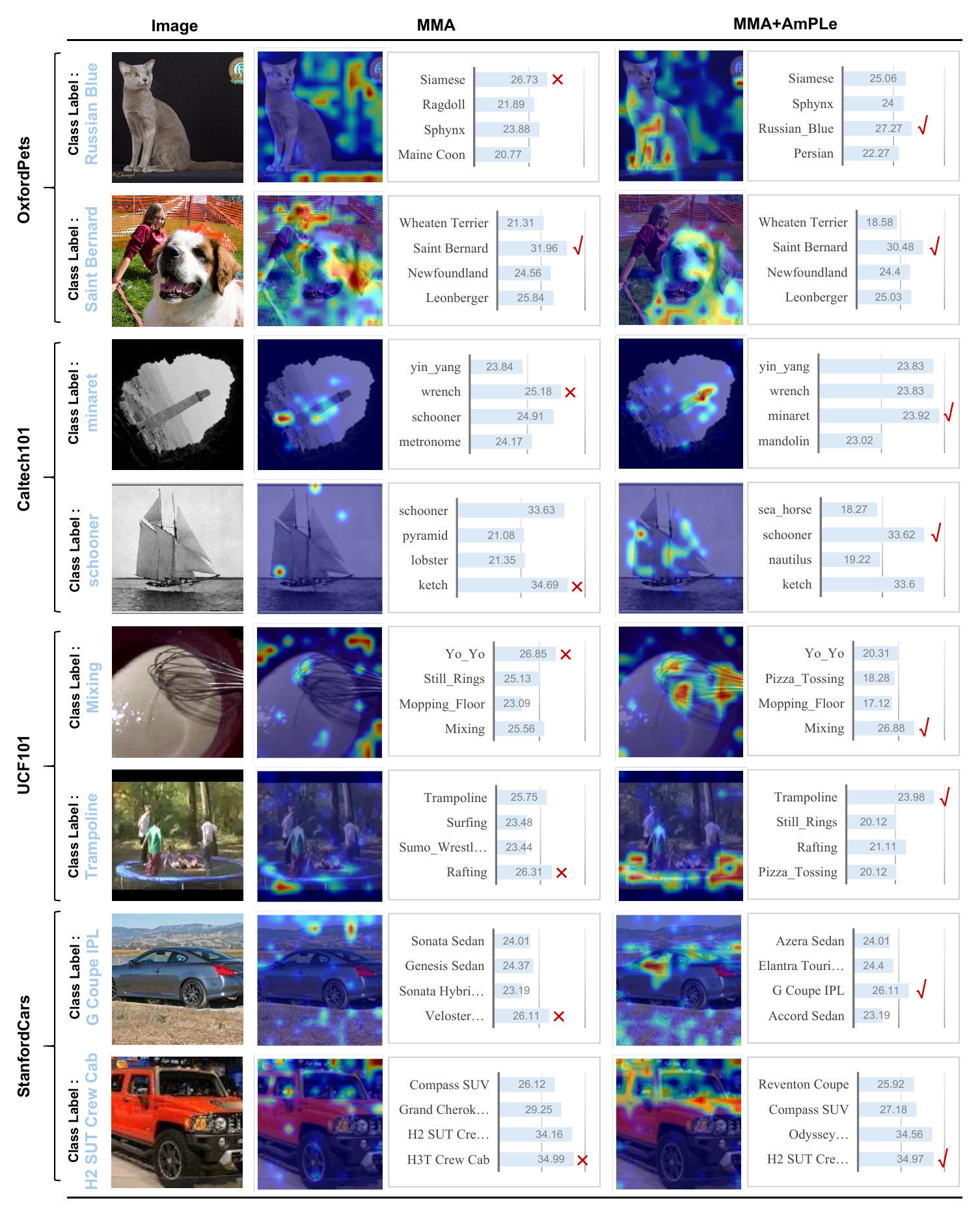}
    \caption{Visual explanation maps generated by Grad-CAM for representative samples from OxfordPets, Caltech101, UCF101, and StanfordCars. The comparisons highlight the differences in attention allocation between MMA and MMA+AmPLe.}
    \label{fig:GradCAM}
\end{figure*}

\label{further_analysis}
\begin{figure*}
    \centering
    \includegraphics[width=0.98\textwidth]{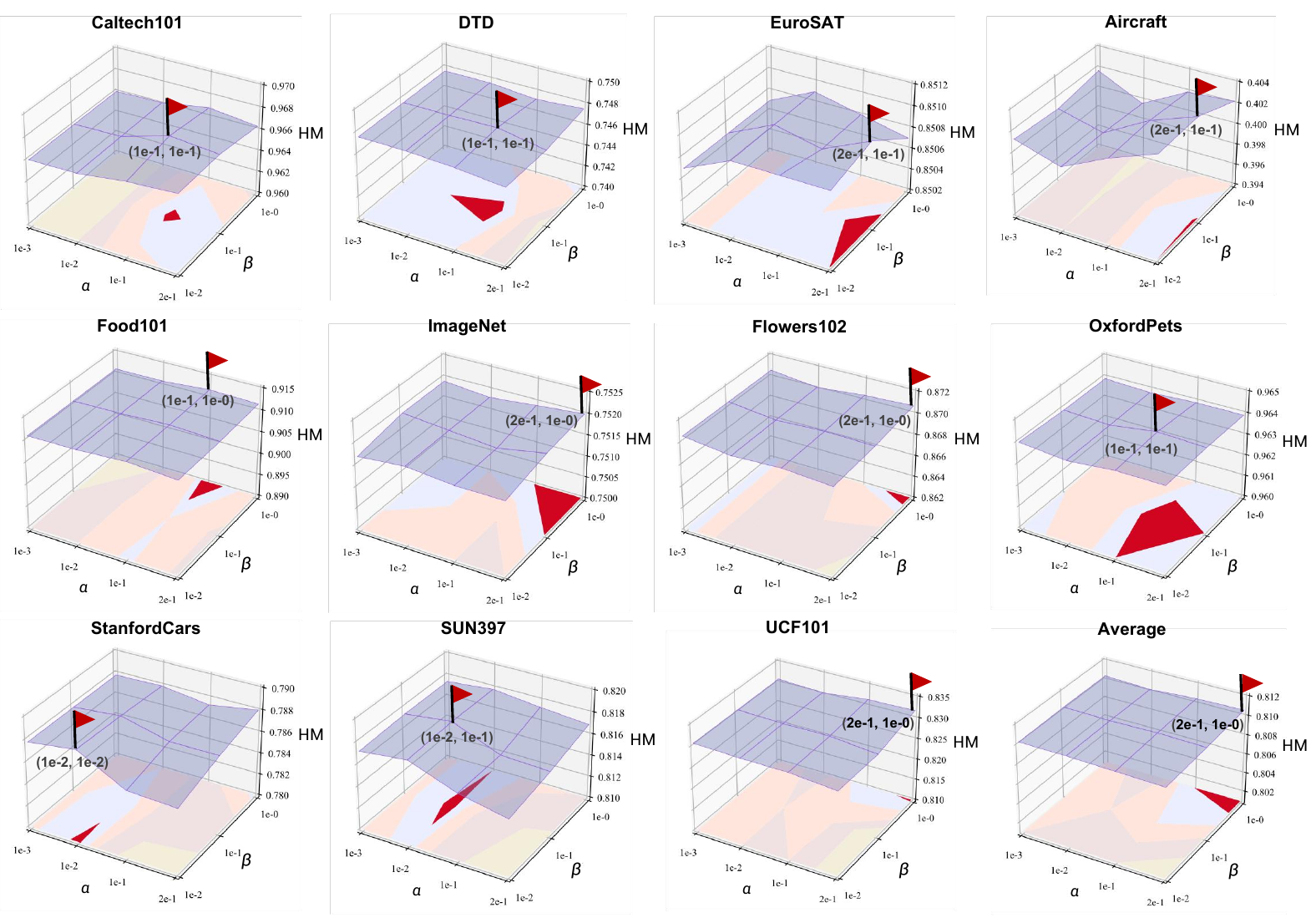}
    \caption{The results of hyperparameter experiments.}
    \label{fig:para_analysis}
\end{figure*}

\begin{figure*}
    \centering
    \includegraphics[width=0.98\textwidth]{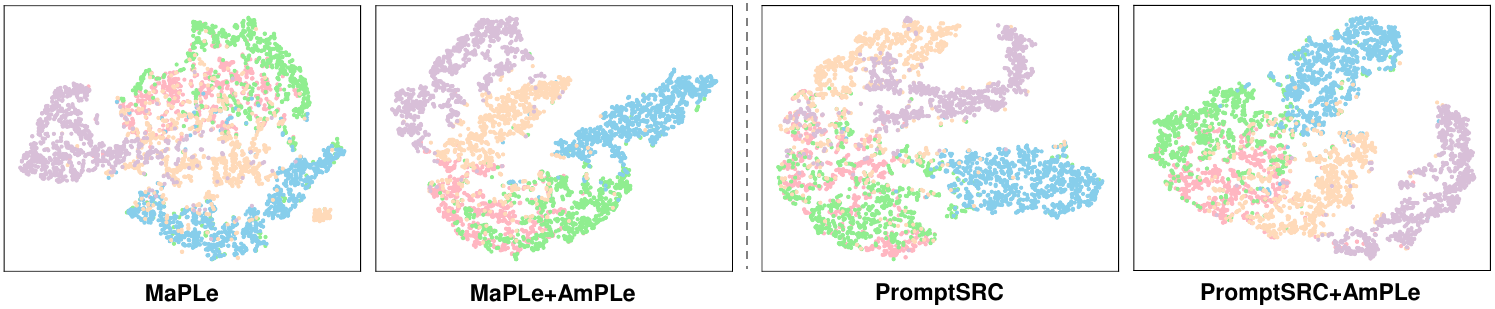}
    \caption{Visual classification results on the EuroSAT dataset in the base-to-novel generalization setting.}
    \label{fig:t-SNE}
\end{figure*}

\textbf{Adaptability to Different Prompt Sets.} To further assess the adaptability of AmPLe to different prompt sets, we conduct additional experiments on the Caltech101 and UCF101 datasets using MMA+AmPLe. Specifically, we considered three types of prompt variations: (1) \textit{Number}: reducing the number of prompts per class; (2) \textit{Structure}: altering the template style (e.g., replacing ``A photo of ... which ...'' with ``An image depicting ... that ...''); and (3) \textit{Content}: generating alternative visual descriptors with GPT-4. The results summarized in Fig.~\ref{fig:prompts_variant} show that prompt variations can indeed affect the model’s performance. We acknowledge this as a limitation. Nevertheless, we can also observe that retraining only the lightweight $\mathcal{W}$ is sufficient to restore or even improve the original results. This finding highlights AmPLe’s practical adaptability, allowing efficient adjustment to new prompt sets with minimal additional cost.

\textbf{Case Study.} \label{case_study}
We employ Grad-CAM~\citep{DBLP:conf/iccv/SelvarajuCDVPB17} to generate visual explanation maps and provide an in-depth analysis of the decision-making behavior of different models. As shown in Fig.~\ref{fig:GradCAM}, we select representative samples from OxfordPets, Caltech101, UCF101, and StanfordCars to compare the attention distributions of MMA and MMA+AmPLe. The visualizations highlight systematic differences in how the two models allocate attention. For MMA, the highlighted regions often spread across class-irrelevant areas. This dispersed focus indicates that MMA is affected by sample-prompt matching bias, where prompt-irrelevant visual information distorts the association between the input samples and the intended class semantics. In contrast, MMA+AmPLe consistently concentrates on class-relevant regions, such as breed-specific facial and body features in OxfordPets, the object’s shape in Caltech101, the action-defining objects in UCF101, and the distinctive car contours in StanfordCars. For example, in action recognition tasks from UCF101, MMA+AmPLe improves the localization of motion-related objects, which are semantically critical for distinguishing actions. In fine-grained object categories such as StanfordCars, AmPLe enhances focus on discriminative car parts, enhancing inter-class separability in visually similar categories. Notably, in the case of the Saint Bernard, although both models produce correct predictions, the Grad-CAM maps expose a qualitative distinction. MMA distributes attention to both the dog and the surrounding human figure, indicating potential susceptibility to background noise. By contrast, MMA+AmPLe places its focus more sharply on the dog’s head and body contours, providing a cleaner alignment between visual evidence and the class label. Overall, these focused attention patterns provide visual evidence that AmPLe mitigates the impact of prompt-irrelevant semantics and strengthens discriminative localization, leading to more robust and interpretable predictions.

\textbf{Hyperparameters Research.} In our proposed AmPLe framework, two key hyperparameters, \(\alpha\) and \(\beta\) (as defined in Equation (\ref{eq:L_final})), influence the models' performance. To systematically analyze their impact, we conduct an empirical evaluation by testing various combinations of \(\alpha\) and \(\beta\). Specifically, we search for the optimal \(\alpha\) within the set \(\{2e-1, 1e-1, 1e-2, 1e-3\}\) and the optimal \(\beta\) within \(\{1e-0, 1e-1, 1e-2\}\) across all 11 datasets under the base-to-novel generalization setting. The values of \(\alpha\) and \(\beta\) are determined empirically, and the results are visualized in Fig. \ref{fig:para_analysis}. As observed, the optimal combination of \(\alpha\) and \(\beta\) varies across different datasets, highlighted by red triangular flags. For instance, the optimal settings for \(\alpha\) and \(\beta\) in the EuroSAT, Food101, and SUN397 datasets are \(\{2e-1, 1e-1\}\), \(\{1e-1, 1e-0\}\), and \(\{1e-2, 1e-1\}\), respectively. Therefore, the elaborate assignment of \(\alpha\) and \(\beta\) can further enhance the discriminative performance of AmPLe. In practice, when adapting AmPLe to new datasets, we can start with the average best-performing setting \((\alpha=2e-1, \beta=1e-0)\), followed by further refinement using a small validation split to optimize performance.

\textbf{Visual Comparison.} To intuitively illustrate the effectiveness of AmPLe, we visualize the predictions of test samples on the EuroSAT dataset using t-SNE~\citep{van2008visualizing}, as shown in Fig. \ref{fig:t-SNE}. Each color represents a different class, with blue, pink, green, orange, and purple denoting five distinct categories. By analyzing the visualization results, we derive two key observations: (1) Models integrated with AmPLe exhibit more compact intra-class clustering, meaning that samples belonging to the same category are more tightly grouped. This aligns with our motivation to mitigate sample-prompt matching bias. By extracting prompt-relevant features, AmPLe reduces intra-class variance and enforces greater consistency within each class in the learned representation space. (2) Incorporating AmPLe enlarges the inter-class distances, suggesting that the learned representations become more discriminative with improved inter-class separability. This outcome is consistent with our motivation to mitigate the model–prompt matching bias. By aggregating diverse prompt-specific predictions across different VLMs, AmPLe strengthens class-discriminative boundaries, thereby making representations of different categories more distinguishable.
These results confirm that AmPLe effectively aids the baseline methods in learning more informative and class-distinctive representations, thereby enhancing the models' robustness and classification performance.

\section{Conclusion}\label{conclusion}
In this work, we conduct exploratory experiments and reveal the presence of model-prompt matching bias and sample-prompt matching bias in multi-prompt learning. The two biases negatively affect downstream task inference when using VLMs. To jointly mitigate the two types of bias, we propose Adaptive-Debiased Ensemble Multi-Prompt Learning (AmPLe), a novel framework that learns the prompt-relevant semantic information from input samples to adaptively calculate debiased ensemble weights and aggregates the diverse prompt-specific predictions derived by multiple prompts and different VLMs. Extensive experiments demonstrate that AmPLe consistently enhances the generalization ability of VLMs. Furthermore, theoretical validation from a causal perspective substantiates the feasibility and reliability of our approach.

\textbf{Limitations and Future Work.} The current approach to the design of multiple prompts relies on class labels, often overlooking specific image details, leading to domain-relevant semantic prompts that may not fully align with the visual content. Additionally, multiple prompts can introduce redundancy or repetitive information, affecting the effective learning of models. Therefore, in the future work, we intend to focus on exploring more contextually relevant prompts based on image context and optimizing the generated prompts to reduce redundancy.

\section*{Acknowledgements}
This work is supported by National Natural Science Foundation of China No. 62406313, Postdoctoral Fellowship Program of China Postdoctoral Science Foundation, Grant No. YJB20250283.

\section*{Data Availability}
All datasets used in this paper are publicly available, and their access links are provided as follows:

\noindent 1. The ImageNet dataset is available at \url{https://image-net.org/index.php}.

\noindent 2. The Caltech101 dataset is available at \url{https://data.caltech.edu/records/mzrjq-6wc02}.

\noindent 3. The OxfordPets dataset is available at \url{https://www.robots.ox.ac.uk/~vgg/data/pets/data/images.tar.gz}.

\noindent 4. The StanfordCars dataset is available at \url{https://github.com/cyizhuo/Stanford_Cars_dataset}.

\noindent 5. The Flowers102 dataset is available at \url{https://www.robots.ox.ac.uk/~vgg/data/flowers/102/102flowers.tgz}.

\noindent 6. The Food101 dataset is available at \url{https://data.vision.ee.ethz.ch/cvl/datasets_extra/food-101/ }.

\noindent 7. The Aircraft dataset is available at \url{https://www.robots.ox.ac.uk/~vgg/data/fgvc-aircraft/archives/fgvc-aircraft-2013b.tar.gz}.

\noindent 8. The SUN397 dataset is available at \url{https://vision.princeton.edu/projects/2010/SUN/}.

\noindent 9. The DTD dataset is available at \url{https://www.robots.ox.ac.uk/~vgg/data/dtd/download/dtd-r1.0.1.tar.gz}.

\noindent 10. The EuroSAT dataset is available at \url{https://github.com/phelber/eurosat}.

\noindent 11. The UCF101 dataset is available at \url{https://drive.google.com/file/d/10Jqome3vtUA2keJkNanAiFpgbyC9Hc2O/view}.

\noindent 12. The ImageNetV2 dataset is available at \url{https://imagenetv2.org/}.

\noindent 13. The ImageNet-Sketch dataset is available at \url{https://github.com/HaohanWang/ImageNet-Sketch}.

\noindent 14. The ImageNet-A dataset is available at \url{https://github.com/hendrycks/natural-adv-examples}.

\noindent 15. The ImageNet-R dataset is available at \url{ https://github.com/hendrycks/imagenet-r}.

\bibliographystyle{sn-basic}

\bibliography{sn-bibliography}

\end{sloppypar}
\end{document}